\documentclass[pdflatex,sn-basic]{sn-jnl}

\jyear{2022}%

\theoremstyle{thmstyleone}%
\usepackage{physics}
\usepackage{xcolor}
\usepackage{comment}

\DeclareMathOperator{\Pow}{Pow}
\newcommand{\anomaly}{\emph{Anomaly}}
\newcommand{\context}{\emph{Context}}

\theoremstyle{thmstyletwo}%

\theoremstyle{thmstylethree}%

\begin{document}

\title[Explainable Contextual Anomaly Detection]{Explainable Contextual Anomaly Detection using Quantile Regression Forests \footnote{Manuscript accepted by \textit{Data Mining and Knowledge Discovery} journal for publication (June 2023). This is a preprint version.}}


\author*{\fnm{Zhong} \sur{Li}}\email{z.li@liacs.leidenuniv.nl}

\author{\fnm{Matthijs} \sur{van Leeuwen}}\email{
m.van.leeuwen@liacs.leidenuniv.nl}

\affil{\orgdiv{Leiden Institute of Advanced Computer Science (LIACS)}, \orgname{Leiden University}, \country{the Netherlands}}


\abstract{Traditional anomaly detection methods aim to identify objects that deviate from most other objects by treating all features equally. In contrast, contextual anomaly detection methods aim to detect objects that deviate from other objects within a \emph{context} of similar objects by dividing the features into contextual features and behavioral features. In this paper, we develop connections between dependency-based traditional anomaly detection methods and contextual anomaly detection methods. Based on resulting insights, we propose a novel approach to inherently interpretable contextual anomaly detection that uses Quantile Regression Forests to model dependencies between features. Extensive experiments on various synthetic and real-world datasets demonstrate that our method outperforms state-of-the-art anomaly detection methods in identifying contextual anomalies in terms of accuracy and interpretability.}

\keywords{Anomaly detection, Anomaly explanation, Outlier detection, Contextual anomaly detection, Quantile regression forests }



\maketitle

\section{Introduction}
\label{sec:introduction}

According to the well-known definition of \cite{hawkins1980identification}, an anomaly\footnote{Given the fact that ``outlier" is often used as a synonym for ``anomaly" in the anomaly detection literature, we will use them interchangeably in this paper.} is an object that is notably different from most remaining objects. \cite{chandola2009anomaly} subdivided anomalies into three types: point anomalies (an object is considered anomalous when compared against the rest of objects), contextual anomalies (an object is anomalous in a specific context), and collective anomalies (a collection of objects is anomalous with respect to the entire dataset). The analysis of anomalies has a wide range of applications, such as in network security  \citep{ahmed2016survey1}, bioinformatics \citep{spinosa2005support}, fraud detection \citep{ahmed2016survey2}, and fault detection and isolation \citep{hwang2009survey}.

Anomaly analysis consists of two equally important tasks: anomaly detection and anomaly explanation. A wealth of `shallow' machine learning based methods, i.e., not based on deep learning, have been proposed to detect anomalies \citep{chandola2009anomaly}. More recently, many deep learning based anomaly detection methods have also been developed \citep{pang2021deep}. 
However, deep learning based anomaly detection methods are notoriously known as not being interpretable, in the sense that generally both the model itself is non-transparent and the resulting anomaly scores are challenging to interpret without the use of a post-hoc explainer. In this paper it is especially the latter that we consider to be problematic, as post-hoc explanations often rely not only on the model but also on the specific explainer used. In addition, deep learning methods typically require large amounts of data and training the models is a time-consuming process. In many real-world applications, however, `native' interpretability (i.e., without posthoc explainer) may be required, and limited data and/or computation time may be available. For these reasons, in this paper we restrict our focus to `shallow' machine learning based methods. In correspondence with this choice, we focus on settings where the amount of data is smaller than is typically required to learn accurate deep models. Most existing shallow methods only consider point anomaly detection, largely ignoring contextual anomaly detection. Moreover, anomaly explanation has received very limited attention. In this paper, we address both the problem of contextual anomaly detection and that of anomaly explanation, for small to moderately sized tabular data having categorical and/or quantitative features.

\begin{figure}[tb]
\centering
\includegraphics[width=12cm]{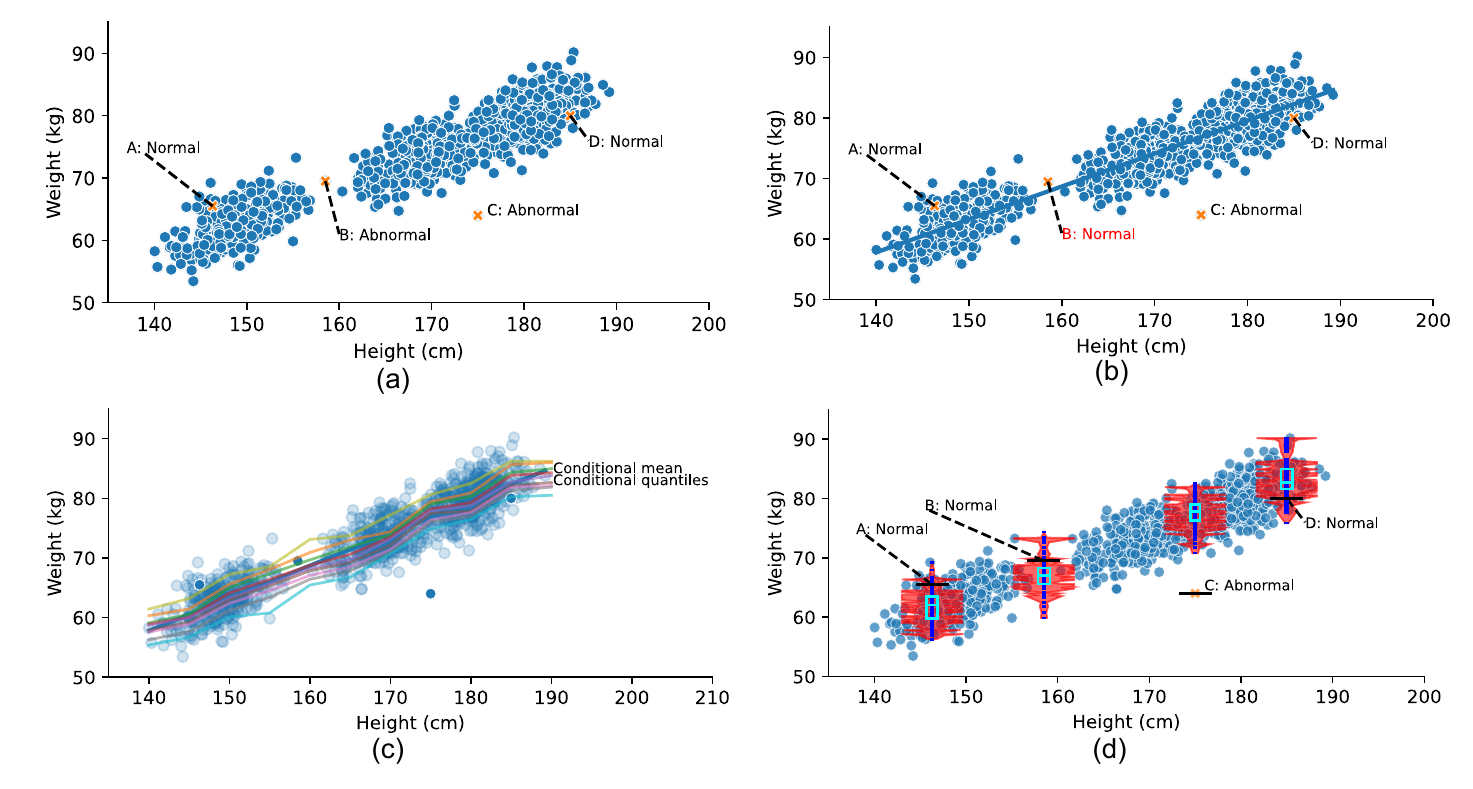}
\caption{Different types of anomaly detection. (a) Distance- and density-based methods both consider object B to be abnormal, because it is far away from other objects and in a low-density region. (b) Dependency-based methods model the relationship between height and weight, and consequently consider object B to be normal. (c) Conditional quantiles provide more information about the conditional distribution than just the mean, and (d) can be used to visualise---using beanplots (or, rather, a variation of a beanplot combined with boxplots, see Section \ref{qcad:ae} for details)---why a certain object is considered (ab)normal.}\label{fig:intro}
\end{figure}

Shallow anomaly detectors are typically categorised into distance-based, density-based, and distribution-based approaches \citep{wang2019progress}. Distance-based and density-based methods use knowledge about the spatial proximity of objects to identify anomalies, while distribution-based methods use knowledge about the distribution of the data to detect anomalies. These methods work under the assumption that \textit{objects having a large distance or different density from their spatial neighbours are anomalous} or \textit{objects that take rare values under a marginal or joint distribution of features are anomalous}, respectively. Using either of these assumptions may lead to false positives though. For example, as shown in Figure~\ref{fig:intro}(a), these methods will mistakenly identify object B as an anomaly if the aim is to detect people that are over- or underweight.

As this is often undesirable, we aim to define and detect anomalies from another perspective, that is, we assume that \textit{objects that violate a dependency, i.e., a relationship between features, are anomalous}. This is the idea behind dependency-based anomaly detection \citep{lu2020dependency}, which is not new but has received limited attention compared to other types of anomalies. It leverages the intrinsic structure and properties of the data to find potentially more relevant anomalies. For example, Figure \ref{fig:intro}(b) shows that this approach will identify object B as normal by modelling the dependency between height and weight.

Traditional anomaly detection techniques---including distance-, density-, and dependency-based methods---treat all features equally when identifying anomalies. However, in domains such as healthcare, sensor networks, and environment protection, some features should never be used directly to quantify anomalousness. For instance, forest fire detection systems should not treat `deviating' values of latitude, longitude, date, and time as an indication of an anomaly. We should not simply discard these features either though, as they may contain relevant information. For example, the `normal' temperature may be higher for certain regions than for others. This motivates us to investigate \emph{contextual} anomaly detection, which can take such extra features into account.

Contextual anomaly detection assumes that \textit{an object is anomalous if it strongly deviates from objects within its `context' of similar objects}. The apparent contradiction in this assumption is explained by the division of the features into two disjoint subsets, i.e., contextual features and behavioral features. The contextual features are only used to define the contexts, using some similarity measure, while the behavioral features are only used to determine whether an object deviates from other objects within its context. Domain knowledge often leads to a natural division between contextual and behavioral features.

We observe that both contextual and dependency-based anomaly detection methods identify anomalies by explicitly or implicitly exploring dependencies between features. Concretely, dependency-based anomaly detection methods model dependency relationships between all features explicitly, while contextual anomaly detection methods model dependency relationships between behavioral and contextual features implicitly or explicitly. As far as we know this connection has not yet been pointed out in the literature. 

\smallskip \noindent
\textbf{Approach and contributions}. 

In this paper, we introduce an approach for contextual anomaly detection and explanation that integrates the core principle of dependency-based anomaly detection into contextual anomaly detection to obtain a very accurate approach. As is common, we use regression analysis to model dependencies. Existing methods for dependency-based and contextual anomaly detection that use regression, however, typically only estimate the conditional mean of the response variable and directly interpret that as `normal' value. This strongly limits how well anomalies can be detected, as the conditional mean provides very limited information about the conditional distribution. We therefore use \emph{quantile regression}, which can model a conditional distribution in much more detail by estimating conditional quantiles. Figure \ref{fig:intro}(c) shows how conditional quantiles provide more information about the relationship between weight and height than a conditional mean could. 

More specifically, a subset of the features, dubbed contextual features, are used to define the context of an object, while the remaining features, dubbed behavioral features, are used for detecting deviations within a context. In this paper we assume that the contextual features can be mixed but all behavioral features are numerical. Given this context and our aims, we use Quantile Regression Forests \citep{meinshausen2006quantile} to perform predictions for each behavioral feature and obtain corresponding uncertainty quantifications. By summing the quantified uncertainties (with a wider quantile interval representing a higher level of uncertainty) for all individual behavioral features, we obtain the anomaly score for a data instance. By attributing parts of the anomaly score to individual behavioral features, the approach intrinsically provides explanations in the sense that it can convey to which extent which features contributed to making the instance an anomaly. This offers advantages to post-hoc explanation methods such as SHAP \citep{lundberg2017unified}, which we do not consider `attributable' for the following two reasons. First, the post-hoc explanation may not match the information/rationale used by the model to detect an anomaly \citep{li2022survey}. Second, it has recently been shown \citep{fokkema2022attribution} that attribute-based explanations cannot be both recourse sensitive and robust, which is a good reason to avoid such explainers when possible and use a `native' attribution-based method instead.

As far as we are aware, we are the first to use quantile regression for dependency-based or contextual anomaly detection. Specifically, we choose to employ Quantile Regression Forests for three reasons. First, it can model both linear and non-linear dependency between features. Second, in the paper that introduced the method it was empirically shown to outperform other conditional quantile estimators in most cases. Figure \ref{fig:intro}(d) shows how the estimated conditional quantiles can be used to approximate the conditional probability density at different locations, which we can use to accurately detect anomalies. Moreover, the quantiles are helpful to explain why an object is considered an anomaly without having to explicitly refer to other objects.


The main contributions of our work can be summarized as follows: (1) We identify a connection between dependency-based traditional anomaly detection methods and contextual anomaly detection methods, and exploit this observation to introduce a novel high-level approach to contextual anomaly detection and explanation; (2) We instantiate this generic approach using quantile regression (and Quantile Regression Forests specifically) for anomaly detection and a beanplot-based visualization for anomaly explanation; and (3) We perform extensive experiments on synthetic and real-world datasets to empirically demonstrate the effectiveness, and interpretability of the proposed method when compared to state-of-the-art methods.

The remainder of this paper is organized as follows. Section 2 discusses related work, both in contextual and traditional anomaly detection. Section 3 introduces notation, formalizes the problem, and presents the high-level approach that we propose. Section 4 then describes some technical preliminaries, most notably Quantile Regression Forests. Section 5 introduces QCAD, our proposed method for Quantile-based Contextual Anomaly Detection and Explanation that instantiates the high-level approach. Section 6 empirically compares QCAD to its competitors, and provides a case study that investigates the use of QCAD to find exceptional football players from data. Section 7 concludes the paper.

\section{Related work}
\label{sec:relatedWork}

We first discuss related work on contextual anomaly detection and explanation, and then proceed with the two most closely related types of traditional anomaly detection: dependency-based and subspace-based anomaly detection.

\subsection{Contextual Anomaly Detection and Explanation}

Contextual anomaly detection has received particular attention in spatial data \citep{cai2013spatial}, temporal data \citep{salvador2004learning}, and spatio-temporal data \citep{smets2009discovering}, where spatial and/or temporal features are used to define contexts. These methods are not directly applicable to other domains, where the contexts are defined by other types of features; `generic' contextual anomaly detection has received limited attention in the community.

CAD \citep{song2007conditional} is a seminal work that introduced generic contextual anomaly detection. It assumes a user-specified partition of features into contextual (called `environmental') and behavioral (called `indicator') features, and uses Gaussian Mixture Models to fit the distributions of the contextual and behavioral feature spaces. Dependencies between contextual features and behavioral features are then learned by means of `mapping functions', and an object is considered anomalous if it violates the learned functions. For this to work CAD assumes that both the contextual and behavioral features consist of an unknown number of multiple Gaussian components, which may be a strong assumption in practice. Further, CAD can only handle numerical features and is computationally very expensive. Our proposed method makes no assumptions about the distribution of the features, can deal with mixed contextual features and numerical behavioral features, and we will show empirically that it is computationally more efficient than CAD while achieving a higher detection accuracy.

ROCOD \citep{liang2016robust} is also closely related, and uses local and global models of expected behavior to describe the dependencies between contextual and behavioral features. Concretely, standard regression models such as CART are used to learn global patterns, with contextual features as predictor variables and behavioral features as response variables. Local patterns are computed based on the means of behavioral feature values of an object's neighbours. An object's actual value is compared to the local and global pattern, and the weighted average of these differences forms the anomaly score. As the conditional mean describes only one aspect of a conditional distribution, and is not necessarily the point with the highest probability of occurrence (i.e., the mode). To address this, our method employs quantile regression analysis to estimate conditional quantiles, which provide a much more complete description of the conditional distribution. As a result, our method empirically outperforms ROCOD in terms of accuracy.

With the increasing use of anomaly detection algorithms in safety-critical domains, such as healthcare and manufacturing, the ethical and regulatory obligations to provide explanations for the high-stakes decisions made by these algorithms has become more pronounced \citep{li2022survey}. Existing contextual anomaly detection do not provide such explanations though, and non-trivial modifications would be needed for them to do so. Specifically, CAD \citep{song2007conditional} computes the anomaly score of a data instance by measuring its deviation from the mapping functions that are learned from the majority of data instances. This approach poses a challenge in generating intrinsic explanations, as it does not allow for the attribution of the anomaly score to individual features. Further, ROCOD \citep{liang2016robust}, the other known method for contextual anomaly detection, calculates the weighted average of an instance’s differences to the learned local and global patterns as its anomaly score. The local pattern is obtained using its neighbours, while the global pattern is learned using all instances, making it difficult to associate the anomaly score with individual features.

Despite the existence of numerous methods for explaining anomalies, such as those outlined in recent (survey) papers \citep{panjei2022survey,li2022survey,xu2021beyond}, there has been very limited research on contextual anomaly explanation. Particularly, COIN \citep{liu2018contextual} explains outliers by reporting their outlierness score, the features that contribute to its abnormality, and a contextual description of its neighbourhoods. COIN treats all features equally though, while our method divides features into contextual and behavioral features. Further, we develop a visualisation that helps explain contextual anomalies in addition to reporting the overall anomaly score, feature importance, contextual neighbours, and individual anomaly score for each behavioral feature.

The following methods are less relevant because they consider slightly different problems. \cite{valko2011conditional} construct a non-parametric graph-based method for conditional anomaly detection, which only addresses the problem of a single categorical behavioral feature. \cite{hayes2014contextual} first identify anomalies on behavioral features, and then refine the detected anomalies by clustering all objects on contextual features. \cite{tang2015mining} detect group anomalies from multidimensional categorical data.  \cite{hong2015multivariate} present a contextual  anomaly detection framework dedicated to categorical behavioral features which also considers the dependencies between behavioral features. Moreover, \cite{zheng2017contextual} apply robust metric learning on contextual features to find more meaningful contextual neighbours and then leverage $k$-NN kernel regression to predict the behavioral feature values. \cite{meghanath2018conout} develop ConOut to automatically find and incorporate multiple contexts to identify and interpret outliers.

\subsection{Traditional Anomaly Detection}

Although traditional anomaly detection considers a problem that is different from contextual anomaly detection, dependency-based and subspace-based anomaly detection leverage techniques that are related to our method.

\subsubsection{Dependency-based Anomaly Detection}
Dependency-based anomaly detection aims to identify anomalies by exploring dependencies between features. \cite{teng1999correcting} explores the dependency between non-target and target features to identify and correct possible noisy data points. To detect networking intrusions, \cite{huang2003cross} present Cross-Feature Analysis to capture the dependency patterns between features in normal networking traffic. To detect disease outbreaks, \cite{wong2003bayesian} propose to explore the dependency between features using a Bayesian network. \cite{noto2010anomaly} propose to detect anomalies by using an ensemble of models, with each model exploring the dependency between a response feature and other features. \cite{babbar2012mining} use Linear Gaussian Bayesian networks to detect anomalies that violate causal relationships. 

LoPAD \citep{lu2020lopad} first uses a Bayesian network to find the Markov Blankets of each feature. Then, a predictive model (e.g., CART) is learnt for each individual feature as response variable, with its Markov Blankets as predictor variables. Given an object, LoPAD computes the Euclidean distance between its actual value and predicted value (i.e., conditional mean) as its anomaly score, for each feature. The resulting anomaly scores are normalized and summed to obtain the final anomaly score for an object. The method does not distinguish contextual and behavioral features and---like ROCOD---uses conditional means to represent the conditional distribution. Consequently, LoPAD cannot (accurately) detect contextual anomalies.

\subsubsection{Subspace-based Anomaly Detection}
Subspace-based anomaly detection seeks to find anomalies in part of the feature space. Specifically, \cite{kriegel2009outlier} propose SOD to identify outliers in varying subspaces. Concretely, they construct axis-parallel subspaces spanned by the neighbours of a given object. On this basis, they investigate whether this object deviates significantly from its neighbours on any of these subspaces. Furthermore, \cite{kriegel2012outlier} extend this work to determine whether an object is anomalous on arbitrarily oriented subspaces spanned by its neighbours. \cite{nguyen2013cmi} propose to find subspaces with strong mutual correlations and then identify anomalies on these subspaces. Finally, \cite{cabero2021archetype} use archetype analysis to project the feature space into various subspaces with linear correlations based on nearest neighbours. On this basis, they explore outliers by ensembling the results obtained on relevant subspaces. Overall, these methods pursue to identify anomalies in a subset of features, but treat all features equally and are thus not suitable to identify contextual anomalies.

\section{Contextual Anomaly Detection and Explanation}
\label{sec:framework}

We first introduce the necessary terminology and notations, and illustrate this with the running example depicted in Table~\ref{tab:ExampleData}. A dataset $\mathbf{X} =\{\mathbf{x}_{1},\ldots,\mathbf{x}_{i},\ldots,\mathbf{x}_{N}\}$ contains $N$ instances (or data points) over the set of features (a.k.a. attributes or variables) denoted by $\mathbf{F} = \{\mathbf{f}^{1},\ldots,\mathbf{f}^{j},\ldots,\mathbf{f}^{D}\}$. $x_{i}^{j}$ denotes the value of the $i$-th object, $\mathbf{x}_{i}$, for the $j$-th feature, $\mathbf{f}^{j}$. In the running example,  we have $N=16$, $D=6$, $\mathbf{F}= \{Latitude, Longitude, Season, Temperature, Rain, Wind\}$, and $x_{1}^{Rain} = 69$.

\begin{table}[!htbp]
\caption{Running example: fictional climate data for Dutch cities. Temperature is measured in degrees Celsius, rain in mm, and wind in miles per hour. The city is not used for anomaly detection. \emph{Latitude}, \emph{Longitude} and \emph{Season} are treated as contextual features, while \textbf{Temperature}, \textbf{Rain} and \textbf{Wind} are considered the behavioral features.}\label{tab:ExampleData}%
\begin{center}
\begin{tabular}{ccccccc}
\toprule
City & \emph{Latitude} & \emph{Longitude} & \emph{Season} & \textbf{Temperature} & \textbf{Rain} & \textbf{Wind}\\
\midrule
Leiden & 52.16 & 4.49 & Winter & 3.0 & 69 & 16\\
Amsterdam & 52.37 &  4.89 & Winter & 2.9 & 55 & 17\\
Rotterdam & 51.92 &  4.46 & Winter & 2.7 & 60 & 15\\
Oss & 51.45 &  5.31 & Winter & 1.1 & 58 & 25\\
Eindhoven & 51.44 &  5.46 & Autumn & 16.6 & 62 & 25\\
Delft & 52.00 & 4.21 & Autumn & 16.1 & 45 & 19\\
Utrecht & 52.09 &  5.10 & Autumn & 18.3 & 42 & 20\\
The Hague & 52.07 &   4.28 & Autumn & 18.5 & 49 & 18\\
Tilburg & 51.33 &  5.52 & Summer & 22.1 & 39 & 22\\
Middelburg & 51.49 &  3.61 & Summer & 20.3 & 41 & 23\\
Arnhem & 51.98 & 5.89 & Summer & 19.6 & 48 & 17\\
Venlo & 51.37 & 6.17 & Summer & 21.8 & 43 & 35\\
Emmen & 52.46 &  6.55 & Spring & 8.3 & 80 & 10\\
Meppel & 52.69 &  6.19  & Spring & 7.1 & 27 & 13\\
Groningen & 53.13 &  6.34 & Spring & 4.2 & 17 & 19\\
Leeuwarden & 53.10 &  5.80 & Spring & 7.2 & 17 & 19\\
\botrule
\end{tabular}
\end{center}
\end{table}

We assume that feature set $\mathbf{F}$ is divided into two disjoint feature sets (typically using domain knowledge): a contextual feature set $\mathbf{C} =\{\mathbf{c}^{1},...,\mathbf{c}^{p},\ldots,\mathbf{c}^{P}\}$ and a behavioral feature set $\mathbf{B} =\{\mathbf{b}^{1},\ldots,\mathbf{b}^{q},\ldots,\mathbf{b}^{Q}\}$, such that $D=P+Q$, $\mathbf{F} = \mathbf{C} \cup \mathbf{B}$, and $\mathbf{C} \cap \mathbf{B} = \emptyset$. Without loss of generality, we can rearrange the features of an object
$\mathbf{x}_{i}=(x_{i}^{1},...,x_{i}^{j},...,x_{i}^{D})$ and represent it as $\mathbf{x}_{i}=(x_{i}^{1},...,x_{i}^{P},x_{i}^{P+1},...,x_{i}^{P+Q}) = (c_{i}^{1},...,c_{i}^{p},...,c_{i}^{P},b_{i}^{1},...,b_{i}^{q},...,b_{i}^{Q})=(\mathbf{c}_{i}, \mathbf{b}_{i}) $, so that $\mathbf{c}_{i}$ and $\mathbf{b}_{i}$ denote its contextual and behavioral feature values, respectively. Accordingly, we refer to the space spanned by $\mathbf{C}$ as \textit{contextual space}, i.e., $\mathcal{C} = \mathbf{c}^{1}\times... \times\mathbf{c}^{p}\times...\times\mathbf{c}^{P}$, and to the space spanned by $\mathbf{B}$ as \textit{behavioral space}, i.e., $\mathcal{B} = \mathbf{b}^{1}\times...\times \mathbf{b}^{q}\times...\times\mathbf{b}^{Q}$.

Finally, let $\Pow$ denote the powerset, i.e., $\Pow(\mathbf{X}) = \{X \subseteq \mathbf{X}\}$.

\subsection{Problem Statement}

In contextual anomaly detection, contextual features are used to determine the so-called \emph{context} of an object. An object's context is used to estimate whether it is anomalous. The latter is achieved by comparing the object's values for the behavioral features to what is `normal' within the object's context---if the object's behavioral values strongly deviate, it is flagged as an anomaly.

For contextual anomaly detection to be meaningful, we must assume that \emph{there exist dependencies between the contextual and behavioral data}. If such a relationship does not exist, there is no need to use contextual anomaly detection; one could simply remove the contextual features and reduce the problem to a traditional anomaly detection problem.

We illustrate this using the running example in Table~\ref{tab:ExampleData}. We can detect anomalous weather taking into account different regions and seasons by specifying contextual feature set $\mathbf{C} = \{Latitude, Longitude, Season\}$ and behavioral feature set  $\mathbf{B} = \{Temperature, Rain, Wind\}$. Each city is now only compared to cities with a similar latitude, longitude, and season, i.e., its context. If a city has values for temperature, rain, and/or wind that strongly deviate from those of the cities in its context, it is marked as anomalous.

For example, Amsterdam and Rotterdam could form the context of Leiden; they are both nearby and we have measurements for the same season. Temperature and wind are similar for all three cities, but there was substantially more rain in Leiden than in Amsterdam and Rotterdam. Hence, for that reason Leiden could be flagged as an anomaly.

To formalise the problem, we introduce a generic `context function' that maps each possible data point to a subset of the dataset, i.e., its context, based on the data point's contextual features.

\textbf{Problem 1: Contextual Anomaly Detection} Given a dataset $\mathbf{X}$ with a feature set $\mathbf{F} = (\mathbf{C},\mathbf{B})$, a context function $\context : \mathcal{C} \rightarrow \Pow(\mathbf{X})$, an anomaly detector $\anomaly : \mathcal{B} \times \Pow(\mathbf{X}) \rightarrow \mathbb{R}^{+}$, and a threshold $\phi$, find all data points for which the anomaly scores exceed $\phi$ and are thereby flagged as anomalous---based on the behavioral features---within their individual contexts, i.e., $\{(\mathbf{c},\mathbf{b}) \in \mathbf{X} \mid \anomaly(\mathbf{b}, \context(\mathbf{c})) \geq \phi \}$.

Note that the context is determined based only on contextual feature values, and that the anomaly detector may only use the behavioral feature values of the data points in the given context when establishing if the given data point is anomalous or not.

In practice, analysts are not only interested in identifying anomalies, but also need to know the underlying reasons for why a specific object is reported as anomaly. This leads to the second problem that we consider.

\textbf{Problem 2: Contextual Anomaly Explanation} Given an anomalous object $\mathbf{x} \in \mathbf{X}$ and the context function $\context$ and anomaly detector $\anomaly$ that were used to detect it, find the behavioral features $\mathbf{B'} \subseteq \mathbf{B}$ for which $\mathbf{x} = (\mathbf{c}, \mathbf{b})$ substantially deviates from $\context(\mathbf{c})$.

\subsection{Overall Approach} \label{subsec:framework}

The problem statement in the previous section suggests a three-pronged approach based on 1) context generation, 2) anomaly detection, and 3) anomaly explanation. In this subsection we explain and illustrate the overall approach that we propose, using the running example from Table~\ref{tab:ExampleData}. 

\begin{figure}[!htbp]
\centering
\includegraphics[width=12cm]{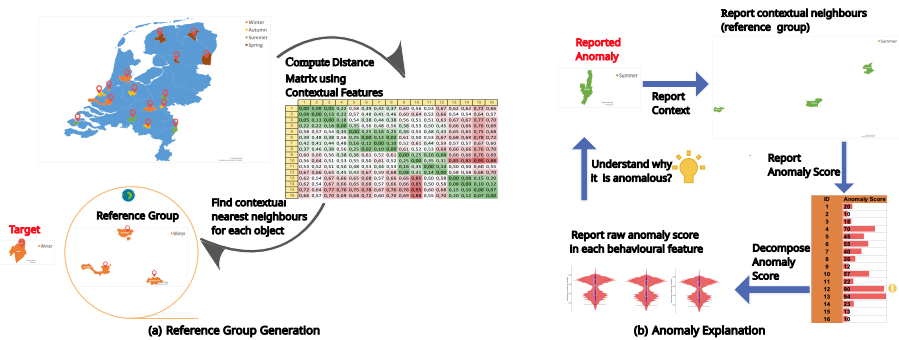}
\caption{ Reference Group Generation and Anomaly Explanation. The Reference Group Generation phase computes the distance matrix using only the contextual features, and finds a reference group for each object on this basis.  The Anomaly Explanation phase produces an explanation for each identified anomaly by reporting its contextual neighbours, final anomaly score, and raw anomaly scores in each behavioral feature.}\label{fig:framework}
\end{figure}

\paragraph{Phase 1: Reference Group Generation} Many choices are possible for the context function; in this manuscript we choose to use an object's $k$ nearest neighbours, which we refer to as \emph{reference group}. The most important reasons for this choice are that 1) once a global contextual distance matrix has been computed the nearest neighbours of any object can be found relatively quickly; 2) this approach only requires a distance metric to be chosen, which can be defined for any type of data and be adapted to the problem at hand; and 3) it is generic enough to allow for different uses of the resulting contexts.

Given a dataset $\mathbf{X}$ with contextual feature set $\mathbf{C}$, we first compute the distance matrix $\mathbf{M}$ between all objects using only the contextual features. Second, for any object $\mathbf{x} \in \mathbf{X}$, we find its $k$ nearest neighbours in \textit{contextual space} based on $\mathbf{M}$. As a result, the $k$ nearest neighbours of $\mathbf{x}$ form a reference group, denoted as $R(\mathbf{x},k)$, which serves as context in Problems 1 and 2.

For instance, as shown in Figure \ref{fig:framework} (a), given  $\{Latitude, Longitude, Season\}$ as the contextual feature set, we first calculate the distance matrix in the contextual space $Latitude \times Longitude \times Season$. Second, based on the distance matrix, we can find the three nearest neighbours of any object as its reference group\footnote{We use $k=3$ for illustrative purposes; in practice, we would have a dataset with far more than 16 records and $k$ would also be much larger.}. Concretely, the three nearest neighbours for $\mathbf{x}_{1}$ are $\{\mathbf{x}_{2},\mathbf{x}_{3},  \mathbf{x}_{4}\}$, which forms a reference group for $\mathbf{x}_{1}$, denoted as $R(\mathbf{x}_{1},3)$.

\paragraph{Phase 2: Anomaly Detection} Given an object $\mathbf{x} \in \mathbf{X}$ with its reference group $R(\mathbf{x},k)$, we apply an anomaly detector $\anomaly$ to obtain an anomaly score \emph{based only on the behavioral attribute values}. We repeat the above process for all objects, leading to $N$ anomaly scores $S= \{s_{1},s_{2},...,s_{N}\}$. We sort $S$ and use threshold $\phi$ to obtain a ranked list of contextual anomalies $A= \{\mathbf{a}_{1},...,\mathbf{a}_{m},...,\mathbf{a}_{M}\}$, with $M \ll N$.

For example, we can apply an anomaly detector on the \textit{behavioral space} $Temperature \times Rain \times Wind$ of $\mathbf{x}_{1}$ and its reference group $R(\mathbf{x}_{1},3)=\{\mathbf{x}_{2},\mathbf{x}_{3},\mathbf{x}_{4}\}$. As a result, we obtain anomaly score $s_{1}$ for $\mathbf{x}_{1}$. Accordingly, repeating this process leads to a set of anomaly scores $S= \{s_{1},...,s_{16}\}$. In our running example we find $\{\mathbf{x}_{4},\mathbf{x}_{12},\mathbf{x}_{13}\}$ as contextual anomalies, as they behave differently in the behavioral space $Temperature \times Rain \times Wind$ when compared to their corresponding neighbours defined in the contextual space $Latitude \times Longitude \times Season$. For example, Oss ($\mathbf{x}_{4}$) has relatively stronger winds in winter when compared to its nearest cities Leiden, Amsterdam and Rotterdam. 

We require anomaly detectors to be \textit{attributable}, meaning that an anomaly score $s$ generated by an anomaly detector must be decomposable into individual contributions towards anomalousness for each behavioral feature using the anomaly detector's native structure.

\paragraph{Phase 3: Anomaly Explanation} For each anomalous object $\mathbf{a}_{m}$, we first report its reference group $R(\mathbf{a}_{m},k)$ using the distance matrix obtained in Phase $1$. Second, we report its anomaly score $s_{m}$, as obtained in Phase $2$. Moreover, we decompose $s_{m}$ into individual contributions from each behavioral feature $\mathbf{b}^{q} \in \mathbf{B}$, resulting in a list of raw anomaly scores $\mathbf{s}_{m}=\{s_{m1},...,s_{mq},...,s_{mQ}\}$. This is possible because  $s_{m}$ is \emph{attributable}. Next, we report the top-$h$ raw anomaly scores in $\mathbf{s}_{m}$ with their corresponding behavioral features, where $h$ can be specified by the analyst. The top-$h$ behavioral features and raw scores enable analysts to better understand why a specific object is flagged as a contextual anomaly.

For example, Figure \ref{fig:framework} (b) reports object $\mathbf{x}_{12}$ as an anomaly. To explain why, we first inspect its reference group $R(\mathbf{x}_{12},3)=\{\mathbf{x}_{10},\mathbf{x}_{11},\mathbf{x}_{12}\}$, which contains the three objects most similar to $\mathbf{x}_{12}$. Second, we report its anomaly score, i.e., $90$. Next, we decompose the score and report the behavioral features with the highest deviations, e.g., $Wind$. We can interpret the result as: $\mathbf{x}_{12}$ deviates substantially in $Wind$ when compared to  $\{\mathbf{x}_{10},\mathbf{x}_{11},\mathbf{x}_{12}\}$, which are all similar in terms of $Latitude$, $Longitude$ and $Season$.

\section{Preliminaries}
\label{sec:prelim}

We first define the conditional mean and conditional quantiles, which we use to motivate the use of conditional quantiles for anomaly detection. Next, we detail quantile regression forests, a model class that can robustly estimate conditional quantiles. Readers familiar with these concepts can skip this section.

\subsection{From Conditional Mean to Conditional Quantiles}
\label{subsec:meantoquantiles}

Given a real-valued variable $V$ and a set of variables $\mathbf{U}$, regression analysis aims to model the distribution of $V$ conditioned on $\mathbf{U}$. Specifically, it takes $V$ as the response variable and $\mathbf{U}$ as the predictor variables to construct a model that can be used to predict $V$ based on $\mathbf{U}$. 
Standard regression uses training data $\{(\mathbf{U}_{1},V_{1}),...,(\mathbf{U}_{n},V_{n})\}$ to learn a model that estimates the conditional mean $E(V \mid \mathbf{U}=\mathbf{u})$ and uses that as prediction for $V$ when $\mathbf{U}=\mathbf{u}$ is given. For example, Least Squares Regression fits a model $\hat{\theta}$ by minimising the expected squared error loss, namely $\hat{\theta} =  \underset{\theta}{\mathrm{argmin}}\, E\{(V-\hat{V}(\theta))^{2})\mid \mathbf{U}=\mathbf{u}\}$. 

The conditional mean, however, is only a single statistic of the conditional distribution and is thereby limited in what it can capture. For example, if the conditional distribution is a multi-modal distribution, the mean is insufficient to describe it (regardless of whether we also consider the standard deviation). 

To allow for more comprehensive descriptions of conditional distributions, \cite{koenker2001quantile} proposed to estimate \emph{conditional quantiles}. As usual, quantiles are splitting points that divide the range of the probability distribution into consecutive intervals having equal probability. Given a continuous variable $V$, its conditional $\alpha$-quantile given $\mathbf{U}=\mathbf{u}$ is defined by $Q_{\alpha}(\mathbf{u}) = \inf \{v: F(v \mid \mathbf{U}=\mathbf{u}) \geq \alpha\}$, where $F(v \mid \mathbf{U}=\mathbf{u}) = P(V \leq v \mid \mathbf{U}=\mathbf{u})$ is the cumulative distribution function. Conditional quantiles have the potential to describe the full conditional distribution of response variable $V$.

Existing regression-based anomaly detection methods typically estimate the conditional mean of $V$ by its expected value  $\hat{v}$, and then use the Euclidean distance between its actual value and the expected value, i.e., $dist(v,\hat{v})$, as anomaly score. It is hard to interpret this distance though, as the shape and range of the conditional distribution are unknown. In this paper we address this by estimating conditional quantiles instead. In particular, we will show that we can use conditional quantiles to approximate the probability of observing a certain data point in its context, which is then used for the anomaly score. 

\subsection{Quantile Regression Forests}
\label{subsec:QRF}

Tree-based regression approaches---such as CART, M5, and Random Forests---are often used to learn both linear and non-linear dependencies. \cite{meinshausen2006quantile} extended the Random Forest to Quantile Regression Forest, which estimates and predicts conditional quantiles instead of means. 

Specifically, a quantile regression forest (QRF) is constructed by building an ensemble of $K$ independent decision trees to estimate the full conditional cumulative distribution function of $V$ given $\mathbf{U} = \mathbf{u}$, based on $n$ independent observations $\{(\mathbf{U}_{1},V_{1}),...,(\mathbf{U}_{i},V_{i}),...,(\mathbf{U}_{n},V_{n})\}$. Each $\mathbf{U}_{i}$ consists of $d$ dimensions. The estimated full conditional distribution can be written as
\begin{equation}
\hat{F}(v \vert \mathbf{U}=\mathbf{u}) = \hat{P}(V \leq v \vert \mathbf{U}=\mathbf{u}) = \hat{E}( \mathbb{I}(V \leq v) \vert \mathbf{U}=\mathbf{u}) = \sum_{i=1}^{n}\omega_{i}(\mathbf{u})\mathbb{I}(V_i \leq v),
\label{eq1}
\end{equation}
where $\omega_{i}(\mathbf{u})$ denotes the weight assigned to observation $(\mathbf{U}_{i},V_{i})$. 

The decision trees that make up a quantile regression forest are constructed similarly to how a random forest is learned, i.e., for each individual tree $m \leq n$ data points are sampled (with replacement), $d^{'} \ll d$ features are randomly selected, and a criterion such as information gain is used to recursively split the data and tree. Each leaf node keeps all its observations though. $K$ decision trees, namely $T_{1}(\theta),...,T_{K}(\theta)$, are independently grown to form a forest.

Once the forest has been constructed, for a given $\mathbf{U} = \mathbf{u}$ each decision tree $T_{j}(\theta)$ is traversed to find the leaf node that $\mathbf{u}$ resides in. A weight $\omega_{i}(\mathbf{u},T_{j}(\theta))$ is then computed 
for each observation $\mathbf{U}_{i}$, with $i \in \{1,...,n\}$: if observation $\mathbf{U}_{i}$ and $\mathbf{u}$ reside in the same leaf node, then $\omega_{i}(\mathbf{u},T_{j}(\theta))$ is defined as 1 divided by the number of samples residing in the leaf node. Otherwise,  $\omega_{i}(\mathbf{u},T_{j}(\theta))$ is $0$. Next, it takes the average of $\omega_{i}(\mathbf{u},T_{j}(\theta))$ over all decision trees, i.e.,
\begin{equation}
\omega_{i}(\mathbf{u}) = \frac{1}{K} \sum_{j=1}^{K}\omega_{i}(\mathbf{u},T_{j}(\theta)),
\label{eq2}
\end{equation}
which is the weight assigned to an observation $\mathbf{U}_{i}$. Finally, conditional quantile $Q_{\alpha}(\mathbf{u}) = \inf \{v: F(v \mid \mathbf{U}=\mathbf{u}) \geq \alpha\}$ can be estimated by $\hat{Q}_{\alpha}(\mathbf{u}) = \inf \{v: \hat{F}(v \mid \mathbf{U}=\mathbf{u}) \geq \alpha\}$. Under reasonable assumptions, \cite{meinshausen2006quantile} proved quantile regression forests to be consistent, i.e., 
\begin{equation}
\mathscr{sup}\abs{\hat{F}(v \mid \mathbf{U}=\mathbf{u}) - F(v \mid \mathbf{U}=\mathbf{u})} \stackrel{p}\longrightarrow 0, \textrm{with } n \longrightarrow \infty
\label{eq3}
\end{equation}
holds pointwise for every $\mathbf{u}$. 

\section{Quantile-based Contextual Anomaly Detection and Explanation}
\label{sec:qcad}

We present an instance of the generic approach for contextual anomaly detection presented in Section~\ref{subsec:framework} that is based on quantile regression forests. 

The main idea of our method is to estimate the deviation of an object’s behavioral values within a given context using uncertainty quantification around predictions, where the predictions are assumed to capture `normal' behavior. Then, a higher uncertainty implies a higher deviation within the context and thus a higher degree of anomalousness. To this end several approaches could be explored. For example, one might use multi-target regression models---such as Multivariate Random Forests \citep{segal2011multivariate}---on all behavioral features, or single-target regression models---such as Random Forests \citep{breiman2001random}---on individual behavioral features followed by aggregation and conformal inference \citep{lei2018distribution}. In this paper we instead opt to use Quantile Regression Forests \citep{meinshausen2006quantile}, a single-target regression model, because it is (relatively) simple, offering advantages with regard to interpretability, and directly provides uncertainty quantifications. More concretely, we derive intervals for the behavioral features from the underlying quantile regression forests as inherent uncertainty quantifications around predictions, resulting in statistically sound and interpretable measures of degree of anomaly.

In the first phase, we generate reference groups using a distance matrix computed on the \emph{contextual space} of all data points. Specifically, we use Gower's distance \citep{gower1971general} (see Section~\ref{prelim:gower} for more detail), to be able to deal with both quantitative and categorical features, and select the $k$ objects having the smallest distances to an object $\mathbf{x}$ as its reference group $R(\mathbf{x}, k)$.

Next, in the second phase, an anomaly score is computed for each individual data point, based on the values in the \emph{behavioral space} of the data point and its reference group. The algorithm is dubbed QCAD---for Quantile-based Contextual Anomaly Detection---and forms the core of our approach; it is introduced in Subsection~\ref{qcad:ad}.
Finally, Section~\ref{qcad:ae} describes how the found anomalies are explained by decomposing the anomaly score in the third phase.

\begin{algorithm}
\caption{Quantile-based Contextual Anomaly Detection (QCAD)}\label{algo:QCAD}
\begin{algorithmic}[1]
\Require Dataset $\mathbf{X}$; contextual feature set  $\mathbf{C}$; behavioral feature set $\mathbf{B}$; number of behavioral features $Q$; number of nearest neighbours $k$; number of conditional quantiles to estimate $n_q$; number of trees $n_t$; maximum number of features used in a tree $n_f$; minimum number of samples to split a node $n_s$.
\Ensure Anomaly score list $\mathbf{S}$
\Procedure{QCAD}{$\mathbf{X},\mathbf{C},\mathbf{B}, k, n_q, n_t, n_f, n_s$}
\State $\mathbf{S} \Leftarrow \{\}$
\For{$\mathbf{x} \in \mathbf{X}$}
    \State $R(\mathbf{x},k) \Leftarrow GetReferenceGroup (\mathbf{X},\mathbf{C},\mathbf{x},k)$
    \For{$q \in \{1,2,...,Q\}$}
    \State $QRF \Leftarrow LearnQRF(R(\mathbf{x},k),\mathbf{C},\mathbf{b}^{q}, n_q, n_t, n_f, n_s)$
    \State $s(\mathbf{x}\rvert\mathbf{b}^{q}) \Leftarrow$ $AnomalyScore(QRF, \mathbf{C},\mathbf{b}^{q}, \mathbf{x})$
    \EndFor
    \State $s(\mathbf{x}) = \frac{1}{Q}\sum_{q = 1}^{Q}s(\mathbf{x}\rvert\mathbf{b}^{q})$
     \State  $\mathbf{S}.append((\mathbf{x},R(\mathbf{x},k),s(\mathbf{x})))$
\EndFor
\State \Return $\mathbf{S}$
\EndProcedure
\end{algorithmic}
\end{algorithm}

\subsection{Detecting Anomalies with Quantile Regression Forests}
\label{qcad:ad}

Algorithm~\ref{algo:QCAD} outlines the QCAD algorithm, which takes a dataset and a number of hyperparameters as input and outputs a list of all data points together with their reference groups and computed anomaly scores.  Specifically, we assume that the contextual features can be mixed but all  behavioral features are numerical. We will first describe the overall algorithm, and then go into the specifics of the score computation.

\paragraph{Algorithm} 

After initializing the empty score list (Line 2), the algorithm iterates over all data points in dataset $\mathbf{X}$ (Ln 3--11). For each object $\mathbf{x} =(c^{1},...,c^{p},...,c^{P},b^{1},...,b^{q},...b^{Q})=(\mathbf{c},\mathbf{b})$ we first obtain the reference group that was computed in the first phase (Ln 4). We then iterate over all features in behavioral feature set $\mathbf{X}$ in order to compute a partial anomaly score for each behavioral feature (Ln 5--8). These partial anomaly scores are summed to obtain the anomaly score for $\mathbf{x}$ (Ln 9), i.e., we assume the behavioral features to all have equal potential to contribute to the overall anomaly score. After this, the data point, its reference group, and its anomaly score are appended to the anomaly score list (Ln 10). Finally, the anomaly score list is returned as output (Ln 12).

Within the inner \texttt{for} loop, we first learn a quantile regression forest using behavioral feature $\mathbf{b}^{q}$ as response variable and the data point's reference group $R(\mathbf{x}, k)$ as training data (Ln 6). All contextual features $\mathbf{C}$ are used as predictor variables for every constructed QRF. We then use the learned quantile regression forest, the features, and the data point to compute the raw partial anomaly score (Ln 7), which we will motivate and explain in detail next.

\paragraph{QRF-based anomaly score}

As argued in the Introduction and Subsection~\ref{subsec:meantoquantiles}, taking the distance between a data point and a conditional mean as basis for a contextual anomaly score may be too limiting: this only works if all `normal' data points reside close to the mean. Instead, we aim to---conceptually---consider the entire conditional probability density function and use the local density of a given data point as a proxy for anomalousness: the lower the density, the higher the anomaly score. Directly accurately estimating the density function is hard though, especially in areas of low density, which are of particular importance to us.

\begin{figure}[bt]
	\centering
	\includegraphics[width=12cm]{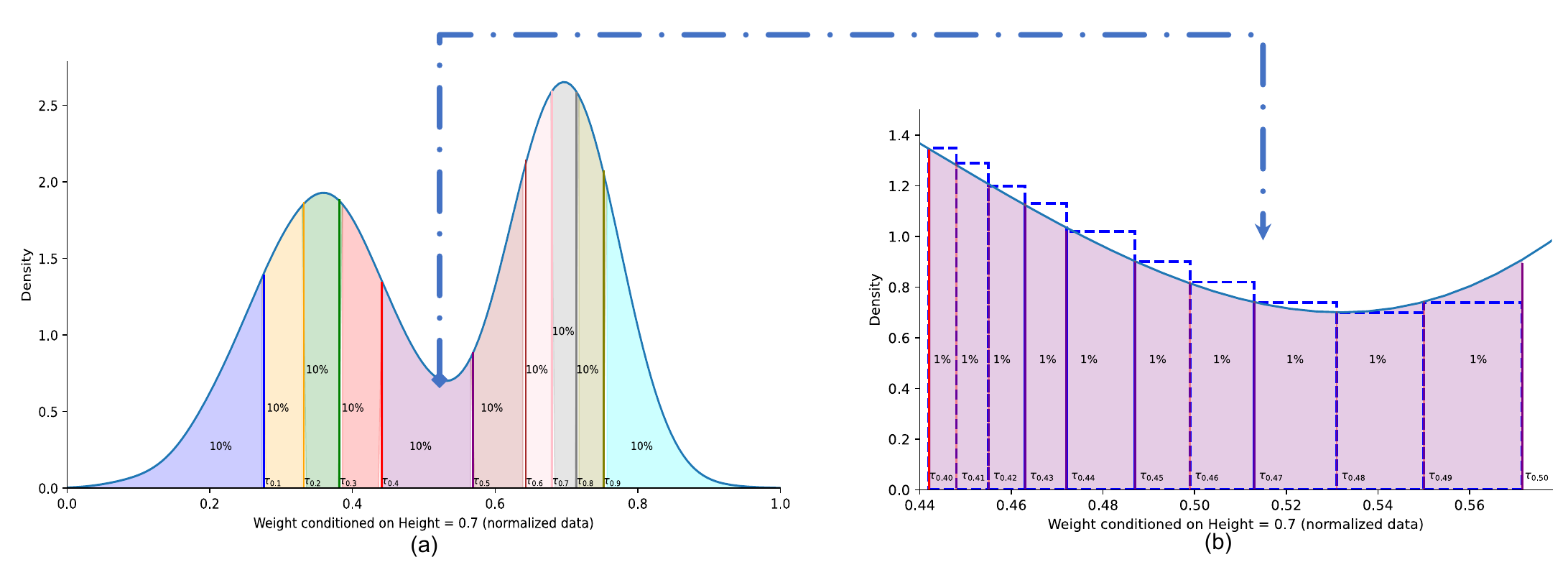}
	\caption{(a) Estimated conditional quantiles $\{Q_{0.1}, Q_{0.2}, \ldots ,Q_{0.9}\}$ for a behavioral feature conditioned on contextual features. (b) Zooming in on the area between $Q_{0.40}$ and $Q_{0.50}$, we see that percentiles are more likely to be  sufficiently detailed than the quantiles in (a).}\label{fig:quantile}
\end{figure}

This is where the quantile regression forests come in: given sufficient training data, they accurately learn the conditional cumulative distribution function, which can be queried in inverse form, i.e., through conditional quantiles. When querying a quantile regression forest, this can be done at different granularities. For example, Figure~\ref{fig:quantile}(a) shows that conditional quantiles $\{Q_{0.1}, Q_{0.2}, \ldots ,Q_{0.9}\}$ may very well be insufficient to accurately describe a conditional distribution as they overly smooth the underlying distribution and thus fail to capture the nuances accurately, while Figure~\ref{fig:quantile}(b) show that conditional \emph{percentiles}, i.e., $\{Q_{0.01}, Q_{0.02}, \ldots ,Q_{0.99}\}$, are much more likely to provide sufficient detail. 

We use conditional percentiles for our anomaly score, because they provide a high level of granularity while not requiring very large amounts of data to be estimated accurately. 
As additional benefit, the difference between each two consecutive percentiles is always assessed by a weighted combination of a comparable number of training data points, i.e., at the cost of some smoothing we do not suffer from extremely poor local density estimates in low density areas.

To formally develop our anomaly score, we first define $\tau_i$ to be the $i$th percentile, i.e., $\tau_i = Q_{i/100}, \forall i \in [0,100]$. For any \emph{percentile interval}, i.e., an interval $[\tau_i, \tau_{i+1}]$ defined by two consecutive percentiles $i$ and $i+1$, we have by definition that it spans exactly $0.01$ probability (see Figure~\ref{fig:quantile}(b)). We could estimate the local density of a percentile interval and use that for our anomaly score, but we aim for a score that becomes larger when a data point is deemed to be more anomalousness. 

To this end we define \emph{percentile interval width} $w_i$ as the difference between two consecutive percentiles $i$ and $i+1$, i.e., $\tau_{i+1} - \tau_i$. As such, interval width can be regarded as `inverse density', meaning that width will increase as the local density decreases. For example, in Figure~\ref{fig:quantile}(b) we have $\tau_{46} = 0.5$ and $\tau_{47} = 0.513$, which gives $w_{46} = 0.013$. Data points that fall in percentile intervals having relatively large widths are more likely to be contextual anomalies, as they reside in low-density areas of the conditional distribution.

The basic idea is thus to define the anomaly score for an object $\mathbf{x}$ and behavioral feature $\mathbf{b}^{q}$ as $w(\mathbf{x}\rvert\mathbf{b}^{q})$, i.e., the width of the (QRF-predicted) percentile interval in which the behavioral value of the data point falls. We need to consider a special case though: the actual behavioral feature value $b^{q}$ may be less than the smallest estimated conditional quantile, i.e., $\tau_{0}^{q}$, or greater than the largest estimated conditional quantile, i.e., $\tau_{100}^{q}$. That is, the actual value may not fall in any estimated conditional percentile interval. To address this we extrapolate beyond $\tau_{0}^{q}$ and $\tau_{100}^{q}$, leading to intermediate anomaly score

\begin{equation}
\centering
is(\mathbf{x}\rvert\mathbf{b}^{q})=\left\{
\begin{aligned}\left(1+\frac{\tau_{0}^{q}-b^{q}}{\tau_{75}^{q}-\tau_{25}^{q}}\right)\mathrm{max}(w(\mathbf{x}\rvert\mathbf{b}^{q}))&, \text{ if } b^{q} < \tau_{0}^{q} ; \\
\left(1+\frac{b^{q}-\tau_{100}^{q}}{\tau_{75}^{q}-\tau_{25}^{q}}\right)\mathrm{max}(w(\mathbf{x}\rvert\mathbf{b}^{q}))&, \text{ if }  b^{q} > \tau_{100}^{q}, \\
w(\mathbf{x}\rvert\mathbf{b}^{q})&, \text{ otherwise},\\
\end{aligned}
\right.
\label{eq4}
\end{equation}
where $\mathrm{max}(w(\mathbf{x}\rvert\mathbf{b}^{q}))$ represents the maximum interval width of all conditional percentile intervals for the behavioral feature $\textbf{b}^{q}$.

Unfortunately, directly using Equation~(\ref{eq4}) as partial anomaly score would make our approach prone to the so-called \textbf{dictator effect}: summing such partial scores for a data point that \emph{strongly} deviates in only a \emph{few} behavioral features would lead to a larger anomaly score than anomalous data points deviating moderately in \emph{many} behavioral features. As a result, data points with few, strong deviations would `dictate' highest scores; this is undesirable.

To avoid the dictator effect, we truncate the partial anomaly scores to a predefined maximum and arrive at the final partial anomaly score as

\begin{equation}
\centering
s(\mathbf{x}\rvert\mathbf{b}^{q})=\left\{
\begin{aligned} \frac{\eta}{100} &, \text{ if } is(\mathbf{x}\rvert\mathbf{b}^{q}) > \frac{\eta}{100}; \\
is(\mathbf{x}\rvert\mathbf{b}^{q})&, \text{ otherwise}, \\
\end{aligned}
\right.
\label{eq6}
\end{equation}

where $\eta$ is a hyperparameter. The rationale for $\eta$ is as follows: if the conditional distribution is uniformly distributed and the behavioral feature has range $[0,1]$, then the expected width of any percentile interval width is $0.01$ and $\eta$ can be interpreted as the maximum number of \emph{expected percentile interval widths}. In the experiments we use $\eta = 10$ because of its strong empirical performance (also shown in the ablation study in Appendix~\ref{Appendix:Ablation}). 

By scaling the behavioral features to $[0,1]$ before anomaly detection, e.g., using min-max normalization, the estimated conditional percentiles should also lie in this interval and the interval widths obtained for the individual behavioral features should thus be comparable. In turn, this implies they can be summed to obtain the final anomaly score for a data point (Algorithm~\ref{algo:QCAD}, Line 9). 

\subsection{Explaining Anomalies with Anomaly Beanplots}
\label{qcad:ae}

In the third phase, we provide explanations for the reported anomalies. From the definition of our anomaly score it is clear that it is \textit{attributable}: the overall score can be decomposed into partial scores for individual behavioral features.

For each identified anomaly $\mathbf{a}$, we report its contextual neighbours $R(\mathbf{a})$ and the final anomaly score as computed on Line~9 of Algorithm~\ref{algo:QCAD}. Further, we report the partial anomaly scores corresponding to the behavioral features, i.e., we report $s(\mathbf{x}\rvert\mathbf{b}^{q})$ for all $q$, ranked from highest to lowest to indicate in which behavioral features the anomaly deviates most. 

\begin{figure}[tb]
	\centering
	\includegraphics[width=9cm]{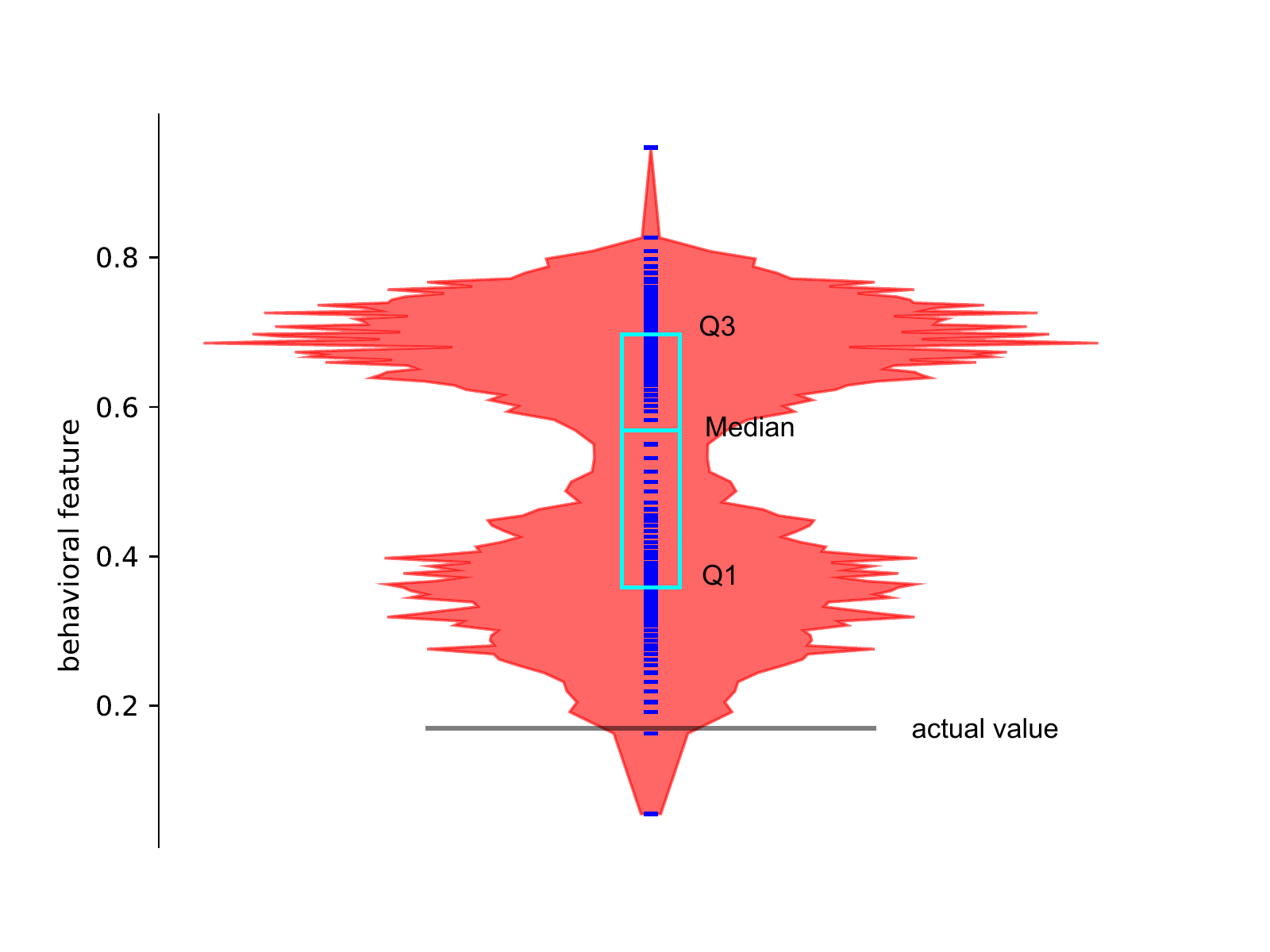}
	\caption{Anomaly beanplot giving insight in what the quantile regression forest learned for a particular data point and behavioral feature, and why the data point is (not) considered an anomaly. The short blue lines indicate the conditional percentiles $\tau_0,\tau_1,\ldots,\tau_{100}$ as learned by the QRF (top to bottom), for the given data point and behavioral feature. As in a box plot, the (cyan) box indicates first quartile ($\tau_{25}$), median ($\tau_{50}$), and third quartile ($\tau_{75}$). The wider red area represents probability densities as estimated based on the conditional percentiles. Finally, the black line indicates the actual value that the data point has.}\label{fig:beanplot}
\end{figure}

Since visualisation often helps to quickly provide valuable insight, we propose the \emph{anomaly beanplot}, a variant of the beanplot by \cite{kampstra2008beanplot}. Figure~\ref{fig:beanplot} shows an example, visually depicting the learned conditional percentiles, their interval widths, and the probability densities that can be estimated from those, all for a particular data point and behavioral feature. (Remember that a quantile regression forest is learned on the reference group of a data point, hence the anomaly beanplot for each data point may be different.)

By including the actual value of the data point in the beanplot (as a horizontal black line), the analyst can easily see how its behavioral feature value is positioned relative to those of its reference group, and why and to what extent it contributes to the its anomalousness. 

\section{Experiments}
\label{sec:experiments}

To demonstrate the effectiveness of our overall approach and proposed method QCAD, we conduct experiments on a wide range of synthetic and real-world datasets. We will first explain the choices regarding datasets, baseline algorithms, and evaluation criteria, after which we present both quantitative results and a case study that investigates interpretability and practical utility.

In addition, we demonstrate the robustness of QCAD with regard to the `number of nearest neighbours' hyperparameter by means of a sensitivity analysis, and conduct several ablation studies to investigate the impact of the hyperparameters. For reasons of space and brevity, the sensitivity analysis and
ablation studies are given in Appendices~\ref{Appendix:Sensitivity} and~\ref{Appendix:Ablation}.

\subsection{Data}
\label{subsec:datasets}

It is challenging to evaluate unsupervised anomaly detection algorithms due to the lack of commonly agreed-upon benchmark data, and down-sampling classification datasets has been criticized for its variation in the nature of the resulting outliers \citep{farber2010using,campos2016evaluation}.

When evaluating unsupervised \emph{contextual} anomaly detection algorithms, this problem is further compounded by the requirement to have both contextual and behavioral features, and---more importantly---treating those differently \citep{liang2016robust}. Consequently, a generally accepted approach is to artificially inject contextual anomalies into existing datasets using a perturbation scheme.

\subsubsection{Data Preprocessing}
\label{prelim:gower}

To make the datasets suitable to all anomaly detection methods, we need to preprocess them before injecting contextual anomalies. First, we leverage Label Encoding \citep{seger2018investigation} to transform categorical contextual features to  numerical form. Second, we employ Min-Max normalisation to scale all behavioral features to $[0,1]$. Min-Max normalization is routinely used in many anomaly detection and generally improves performance \citep{kandanaarachchi2020normalization}.

\textbf{Gower's Distance} To be able to calculate the similarity between two data points containing both categorical and numerical features, we can utilise Gower's distance \citep{gower1971general}. Specifically, the Gower's distance between data points $\mathbf{c}_{i} = (c_{i}^{1},...,c_{i}^{p},...c_{i}^{P})$ and $\mathbf{c}_{j} = (c_{j}^{1},...,c_{j}^{p},...c_{j}^{P})$ is defined as $1- \frac{1}{P}\sum_{p=1}^{P}ps_{ij}^{p}$, where $ps_{ij}^{p}$ represents the partial similarity between data instances $\mathbf{c}_{i}$ and $\mathbf{c}_{j}$ in the $p$-th dimension. For a numerical feature, $ps_{ij}^{p} = 1- \frac{\abs{c_{i}^{p}-c_{j}^{p}}}{\mathrm{max}(\mathbf{c}^p)-\mathrm{min}(\mathbf{c}^p)}$, with $\mathrm{max}(\mathbf{c}^p)$ and $\mathrm{min}(\mathbf{c}^p)$ denoting the maximum and minimum value of all data points for the $p$-th feature, respectively. For a categorical feature, $ps_{ij}^{p} = \mathbb{I}(c_{i}^p-c_{j}^p$), where $\mathbb{I}(\cdot)$ represents the indicator function. Consequently, Gower's distance between two data points is always in $[0,1]$, with a lower value indicating a larger similarity.

\subsubsection{Perturbation Scheme for Outlier Injection}

\cite{song2007conditional} proposed a perturbation scheme to inject contextual anomalies in datasets without ground-truth anomalies, which has become a de-facto standard for the evaluation of contextual anomaly detection~\citep{song2007conditional, liang2016robust, zheng2017contextual, calikus2021wisdom}.



This perturbation scheme, however, has also been criticised for two reasons~\citep{song2007conditional,kuo2018detecting}. 
First, the objects obtained by simply swapping the feature values are still likely to be contextually normal. Second, some very common types of anomalies cannot be yielded through this perturbation scheme. For example, one cannot obtain extreme values by swapping features values in a clean dataset, whereas most anomaly detection methods assume the training dataset is uncontaminated. To avoid these problems, \cite{kuo2018detecting} introduced another perturbation scheme to inject anomalies. We develop a new perturbation scheme by refining this scheme, as follows. 

Given a dataset $\mathbf{X}$ containing $N$ data instances with contextual feature set $\mathbf{C} = \{\mathbf{c}^{1},...,\mathbf{c}^{P}\}$ and behavioral feature set $\mathbf{B} = \{\mathbf{b}^{1},...,\mathbf{b}^{Q}\}$, we first use Min-Max normalization to scale the behavioral features of all objects to $[0,1]$ (keeping the contextual features intact), resulting in a new dataset $\Tilde{\mathbf{X}}$. Second, to inject $m$ anomalies into $\Tilde{\mathbf{X}}$, with $0<m \ll N$, we select $m$ objects $\mathbf{x}_{1},\ldots,\mathbf{x}_{m}$ uniformly at random from $\Tilde{\mathbf{X}}$. For each $\mathbf{x} =(\mathbf{c},\mathbf{b})$ in $\Tilde{\mathbf{X}}$, $\mathbf{c}$ represents the contextual feature values and  $\mathbf{b}$ denotes the behavioral feature values. Third, for a selected object $\mathbf{x}_{i} = (\mathbf{c}_{i}, \mathbf{b}_{i})= (c^{1}_{i},...,c^{p}_{i},...,c^{P}_{i},b^{1}_{i},...,b^{q}_{i},...,b^{Q}_{i})$ and behavioral feature $\mathbf{b}^{q}$, we sample a number uniformly at random from $[-0.5,-0.1] \bigcup [0.1,0.5]$, and then add this random number to the behavioral feature value of $\mathbf{x}_{i}$, namely $b^{q}_{i}$, resulting in $\hat{b}^{q}_{i}$. In the same manner, we repeat this process for each behavioral feature, resulting in $(\hat{b}^{1}_{i},...,\hat{b}^{Q}_{i})$, or $\hat{\mathbf{b}}_{i}$. Accordingly, we generate a new object
$\tilde{\mathbf{x}} = (\mathbf{c}, \hat{\mathbf{b}})$ as contextual anomaly. Fourth, we repeat the third step for each selected object, leading to $m$ perturbed objects.
Fifth and final, we replace the selected objects in the original dataset with their corresponding perturbed objects. To allow extreme values to be injected, we deliberately do not truncate the values outside $[0,1]$ after adding a random number in each behavioral feature.

Our perturbation scheme has the following advantages. When compared to the swapping perturbation scheme proposed by \cite{song2007conditional}, the objects obtained by our perturbation scheme  are very unlikely to remain contextually normal. In addition, our perturbation scheme can---but does not always---lead to extreme values. Note that we do not strictly follow the perturbation scheme proposed by \cite{kuo2018detecting} because their method only adds a non-negative number to the behavioral features. On the one hand, sometimes this non-negative number is zero, leading to injecting a normal object as `anomaly'. On the other hand, sometimes this non-negative number is huge, which makes the injected object (too) easy to detect. Our perturbation scheme avoids these problems by firstly normalising the behavioral feature values, and then setting more reasonable lower and upper bounds for the perturbation. 

\subsubsection{Datasets}

To evaluate and compare our method on a diverse range of datasets, we generate $10$ synthetic datasets and select $20$ real-world datasets; their properties are summarised in Tables~\ref{tab:data_summary1} and \ref{tab:data_summary2} respectively. 

\begin{table}[!htbp]
	\caption{Summary of synthetic datasets. $\#Num, \#Cat, \#\mathbf{C}$ and $\#\mathbf{B}$ represent the number of numerical features, the number of categorical/nominal features, the number of contextual features, and the number of behavioral features, respectively. All behavioral features are numerical, the contextual features can be mixed.}
	\centering
	\begin{tabular}{cccccc}
		\toprule
		Dataset  &Scheme & \#Num    & \#Cat  & \#$\mathbf{C}$  & \#$\mathbf{B}$  \\
		\midrule
		Syn1  & S1 & 8 & 2 & 5 & 5  \\
		\midrule
		Syn2  & S2 & 8 & 2  & 5 & 5  \\
		\midrule
		Syn3  & S3 & 8 & 2  & 5 & 5  \\
		\midrule
		Syn4  & S4 & 8 & 2 & 5 & 5  \\
		\midrule
		Syn5  & S5 & 8 & 2  & 5 & 5  \\
		\midrule
		Syn6  & S1 & 19 & 6 & 20 & 5  \\
		\midrule
		Syn7  & S2 & 19 & 6  & 20 & 5  \\
		\midrule
		Syn8  & S3 & 19 & 6  & 20 & 5  \\
		\midrule
		Syn9  & S4 & 19 & 6  & 20 & 5  \\
		\midrule
		Syn10  & S5 & 19 & 6 & 20 & 5  \\
		\bottomrule
	\end{tabular}
	\label{tab:data_summary1}
\end{table}

We first discuss the synthetic data. To be able to produce data with various forms and degrees of dependencies between behavioral and contextual features, we propose the following generation schemes. For $q \in \{1,...,Q\}$, we have
\begin{itemize}
    \item[(S1)] $\mathbf{b}^{q} = \sum_{p=1}^{Q}\left(\alpha_{qp}\cdot\mathbf{c}^{p}\right)+\boldsymbol{\epsilon}$;
    \item[(S2)] $\mathbf{b}^{q} = \sum_{p=1}^{Q}\left(\alpha_{qp}\cdot(\mathbf{c}^{p})^{3}\right)+\boldsymbol{\epsilon}$,
    \item[(S3)] $\mathbf{b}^{q} = \sum_{p=1}^{Q}\left(\alpha_{qp}\cdot\mathrm{sin}(\mathbf{c}^{p})\right)+\boldsymbol{\epsilon}$,
    \item[(S4)] $\mathbf{b}^{q} = \sum_{p=1}^{Q}\left(\alpha_{qp}\cdot\mathrm{log}(\mathbf{1}+\vert\mathbf{c}^{p}\vert)\right)+\boldsymbol{\epsilon}$,
    \item[(S5)] $\mathbf{b}^{q} = \sum_{p=1}^{Q}\left(\alpha_{qp}\cdot\mathbf{c}^{p}+\beta_{qp}\cdot(\mathbf{c}^{p})^{3}+\gamma_{qp}\cdot\mathrm{sin}(\mathbf{c}^{p})+\delta_{qp}\cdot\mathrm{log}(\mathbf{1}+\vert\mathbf{c}^{p}\vert)\right)+\boldsymbol{\epsilon}$,
\end{itemize}
where $\alpha_{qp}, \beta_{qp},\gamma_{qp}, \delta_{qp} \stackrel{i.i.d.}{\sim} \mathcal{U}(0,1)$ and are replaced by zero with a probability of $1/3$. Further, $\boldsymbol{\epsilon} = (\epsilon_{1},...,\epsilon_{n},...,\epsilon_{N})^{T}$, with $\epsilon_{n} \stackrel{i.i.d.}{\sim} \mathcal{U}(0,0.05)$. 

In addition, for $p \in \{1,...,P\}$, $\mathbf{c}^{p}$ is generated from a Gaussian mixture model with five clusters. If it is numerical, each of the Gaussian centroids is sampled uniformly at random from $[0,1]$ and the diagonal element of the covariance matrix is fixed at $1/4$ of the average pairwise distance between the centroids in each behavioral feature. Otherwise, the centroids are sampled uniformly at random from $\{0,1,...,10\}$ with the same covariance setting. Moreover, every generated number is rounded to an integer to ensure that it is categorical. On this basis, we generate a wide collection of synthetic datasets by varying the generation scheme and the number of contextual features and behavioral features. Sample size is always set to $2000$, and the rate of injected anomalies is fixed at $2.5\%$. See Table~\ref{tab:data_summary1} for an overview.

\begin{table}[!htbp]
	\caption{Summary of real-world datasets. $\#Num, \#Cat, \#\mathbf{C}$ and $\#\mathbf{B}$ represent the number of numerical features, the number of categorical/nominal features, the number of contextual features, and the number of behavioral features, respectively. All behavioral features are numerical, the contextual features can be mixed.}
	\centering
	\begin{tabular}{ccccccc}
		\toprule
		Dataset  & \#Num    & \#Cat  & \#Samples & \#Anomalies (Ratio) & \#$\mathbf{C}$  & \#$\mathbf{B}$  \\
		\midrule
		Abalone & 8 & 1  & 4177 & 100 (2.4\%) & 4 & 5  \\
		AirFoil & 6 & 0  & 1503 & 70 (4.7\%) & 5 & 1  \\
		BodyFat  & 15 &  0  & 252 & 20 (7.9\%)  & 13 & 2   \\
		Boston & 12 &  2  & 506 & 40 (7.9\%)  & 13 & 1   \\
		Concrete & 9 & 0  & 1030 & 50 (4.8\%) & 8 & 1  \\
		ElNino & 3 & 8  & 20000 & 200 (1\%) & 6 & 5 \\
		Energy  & 10 & 0 & 768 & 50 (6.5\%) & 8 & 2  \\	
		FishWeight & 6 & 1  & 157 & 15 (9.5\%) & 6 & 1  \\
		ForestFires  & 11 & 2 & 517 & 50 (9.7\%) & 4 & 9  \\
		GasEmission & 11 & 0 & 7384 & 100 (1.4\%) & 8 & 3  \\
		HeartFailure  & 7 & 5 & 299 & 30 (10\%) & 6 & 6  \\
		Hepatitis  & 11 & 2 & 615 & 30 (4.9\%) & 3 & 10  \\
		LiverPatient  & 9 & 2 & 579 & 30 (5.2\%) & 3 & 8  \\
		Maintenance & 18 & 0 & 11934 & 100 (0.8\%) & 15 & 3  \\
		Parkinson & 21 & 1 & 5875 & 100 (1.7\%) & 20 & 2 \\
        PowerPlant & 5 & 0  & 9568 & 100 (1\%) & 4 & 1  \\
		QSRanking & 12 & 1 & 475 & 40 (8.4\%) & 7 & 6  \\
		SynMachine  & 5 & 0 & 557 & 50 (8.9\%) & 4 & 1  \\
		Toxicity & 7 & 0  & 908 & 50 (5.5\%) & 6 & 1  \\
		Yacht & 7 & 0  & 308 & 30 (9.7\%) & 6 & 1  \\
		\bottomrule
	\end{tabular}
	\label{tab:data_summary2}
\end{table}

Next, we employ the above-mentioned perturbation scheme to inject contextual anomalies into $20$ real-world datasets. We use these datasets because they are representative of the potential application areas of our QCAD framework. That is, they stem from  healthcare \& life sciences (e.g., Bodyfat, Heart Failure, Indian River Patient, Hepatitis, Parkinson Telemonitoring, Abalone, Fish Weight, QSAR Fish Toxicity), social  sciences (e.g., Boston House Price, University Ranking), environmental-protection area (e.g., El Nino, Forest Fires), and engineering (e.g., Gas Turbine CO and NOx Emission,  Yacht Hydrodynamics, Condition Based Maintenance of Naval Propulsion Plants, Synchronous Machine, Airfoil Self-Noise, Concrete Compressive Strength, Combined Cycle Power Plant). A summary is given in Table \ref{tab:data_summary2}; each dataset is described in more detail in Appendix~\ref{Appendix:Data}.

\subsection{Baseline Algorithms and Implementations}
\label{subsec:baselines}

We empirically compare our method to state-of-the-art algorithms, including traditional anomaly detection methods (distance-based, density-based, dependency-based, etc.) and contextual anomaly detection methods. For a fair comparison, we select the following anomaly detectors, which all return an anomaly score---rather than a binary outcome---for each data instance.
\begin{itemize}
    \item Local Prediction Approach to Anomaly Detection (LoPAD) by \cite{lu2020lopad}, which is the state-of-the-art dependency-based traditional anomaly detector;
    \item Conditional Outlier Detection (CAD) by \cite{song2007conditional}, which was the first anomaly detector dedicated to identify contextual anomalies;
    \item Robust Contextual Outlier Detection (ROCOD) by \cite{liang2016robust}, which is the state-of-the-art contextual anomaly detector;
    \item Isolation Forest (IForest) by \cite{liu2008isolation},  which is one of the state-of-the-art isolation-based traditional anomaly detectors;
    \item Local Outlier Factor (LOF) by \cite{breunig2000lof}, which is one of the state-of-the-art density-based traditional anomaly detectors;
    \item $k$-NN anomaly detector by \cite{angiulli2002fast}, which is one of the state-of-the-art distance-based traditional anomaly detectors;
    \item Anomaly detector using axis-parallel subspaces (SOD) by \cite{kriegel2009outlier}, which is one of the state-of-the-art subspace-based traditional anomaly detectors; and
    \item Histogram-Based Outlier Score (HBOS) by \cite{goldstein2012histogram}, which is one of the state-of-the-art histogram-based traditional anomaly detectors.
\end{itemize}

We implemented and ran all algorithms in Python $3.8$ on a computer with Apple M1 chip 8-core CPU and 8GB unified memory. For classical algorithms such as IForest, LOF, $k$-NN, SOD, and HBOS, we use their publicly available implementations in PyOD \citep{zhao2019pyod} with their default settings. Unfortunately, for LoPAD, CAD and ROCOD no implementation was publicly available. For LoPAD, we first use the `bnlearn' package \citep{scutari2019package} in R to find the Markov Blankets of each dataset based on Fast-IAMB \citep{yaramakala2005speculative}. Next, we implement the LoPAD algorithm in Python using CART \citep{breiman2017classification} as the prediction model with bagging size $200$. For CAD, we implement the CAD-GMM-Full algorithm as recommended by \cite{song2007conditional}, with default settings except for parameter `number of Gaussian component'; if this parameter would be set to its default 30, it would take more than one month to finish all the experiments on our computer. In addition, preliminary experiments show that the difference in results when this parameter is set to $30$ and $5$ respectively is negligible for most datasets. We therefore set this parameter to $5$ in all experiments. We implemented ROCOD in Python with its default settings. One important parameter, namely the distance threshold used to find neighbours, is not discussed in \cite{liang2016robust} though. For different datasets and distance metrics, it is hard to obtain a single best value for this parameter and preliminary experiments revealed that ROCOD is sensitive to this parameter. Nevertheless, we also observed that it often achieves relatively good results when the distance threshold used to find neighbours is set to $0.9$; hence, we decided to set this parameter to $0.9$ by default.

\subsubsection{QCAD Parameters Setting}
\label{subsec:ParametersSettings}

As summarised in Table \ref{tab:parameters}, the QCAD algorithm also requires several parameters to be set. First, in our contextual anomaly framework, we need to set the number of nearest neighbours ($k$) used to generate reference group. Second, when creating quantile regression forests, we need to specify the following parameters: the number of trees, the number of maximal features used to construct a tree, and the number of minimal sample size to split in a node of tree. Third, we need to specify the number of conditional quantiles ($l$) to estimate for an object in each behavioral feature. Last, we also need to set the number of features ($h$) used to generate explanations. We set these parameters as follows.

\begin{table}
  \caption{Summary of parameters involved in QCAD. Particularly, \textbf{Value} represents the values that we recommend to use in experiments.}
  \label{tab:parameters}
  \centering
  \begin{tabular}{ccc}
    \toprule
    \textbf{Symbol} & \textbf{Meaning}  & \textbf{Value}\\
    \midrule
    $k$ & number of nearest neighbours & $\mathrm{min}(N/2,500)$\\
    $n_q$ & number of conditional quantiles to estimate & $100$\\    
    $n_t$ & number of trees used to construct a QRF & $100$ or $10$\\    
    $n_f$ & number of maximal features used to construct a tree & $\vert\mathbf{C}\vert$\\  
    $n_s$ & minimum number of samples to split a node  & $10$\\ 
    $h$ & number of features used to generate explanations  & $\mathrm{min}(\vert \mathbf{B} \vert,3)$\\ 
  \bottomrule
\end{tabular}
\end{table}

\begin{itemize}
\item The number of nearest neighbours ($k$): the sensitivity analysis (see appendix) indicates that our approach is robust with respect to this parameter as long as its value is not overly small. By default, we set this parameter to $\mathrm{min}(N/2,500)$, where $N$ is the sample size. 

\item The number of conditional quantiles to estimate ($n_q$): theoretically, an increase in this number will result in better performance in terms of accuracy, at the expense of a larger running time. However, preliminary experiments show that increasing this number beyond $100$ will only produce slightly better results. Therefore, we set it to $100$ by default.

\item The number of trees used to construct a quantile regression forest ($n_t$): in theory, a larger number of trees will produce better performance in terms of accuracy, but at the cost of a larger running time. We empirically found that 100 trees usually gives good results and further increasing the number of trees leads to negligible improvement. Due to time constraints, we set this number to 10 in all experiments in this paper.

\item The number of maximal features used to construct a tree in a quantile regression forest ($n_f$):  \cite{meinshausen2006quantile} demonstrated the stability of quantile regression forest on this parameter, and set this parameter to $1/3$ of the number of variables in their experiments. However, in our experiments, sometimes the number of contextual features is less than 3. To render quantile regression forests applicable on various datasets, we set this parameter to the number of all variables by default. 

\item The minimal sample size for the node of a tree to be split ($n_s$): as indicated in \cite{meinshausen2006quantile}, different values of this parameter do not seem to have much effect on the results, and our preliminary experiments are also in line with this statement. Therefore, we set this number to $10$ by default, as also used in \cite{meinshausen2006quantile}.
\item The number of features used to generate explanations ($h$): This parameter is set to $\mathrm{min}(\vert \mathbf{B} \vert,3)$ by default. However, the end-users can set this parameter according to their preferences as long as its value is between $0$ and $\vert \mathbf{B} \vert $, where 
$\vert \mathbf{B} \vert $ represents the number of behavioral features.
\end{itemize}

For reproducibility, we make all code and datasets publicly available\footnote{https://github.com/ZhongLIFR/QCAD}.

\subsection{Evaluation Criteria}
\label{subsec:metrics}

PRC AUC \citep{liang2016robust,kuo2018detecting}, ROC AUC \citep{micenkova2014learning, micenkova2015learning, pasillas2016unsupervised}, and Precision@$n$ \citep{aggarwal2017outlier} are widely used for the evaluation and comparison of anomaly detection methods that generate a full list of anomaly scores for all observations. They are defined as follows:
\begin{itemize}
    \item Receiver Operating Characteristic (ROC), which is obtained by plotting the true positive rate (y-axis) versus the false positive rate (x-axis) at various threshold settings. The area under this curve, namely ROC AUC, is a threshold-agnostic performance measure widely used in anomaly detection;
    \item Precision-Recall Curve (PRC), which is created by plotting precision (y-axis) against recall (x-axis) at various threshold settings. The area under this curve, namely PRC AUC, is another widely-used, threshold-agnostic performance measure, and is also called Average Precision;
    \item Precision at $n$, or P@n, is defined as the precision of the observations ranked among the top-$n$, where $n\in\{1,2,...,N\}$. In our experiments, we set $n$ to the number of injected contextual anomalies.
\end{itemize}
For completeness: precision is defined as $\frac{\#\{\text{Real anomalies}\} \cap \{\text{Reported anomalies}\}}{\#\{\text{Reported anomalies}\}}$, while recall is defined as $\frac{\#\{\text{Real anomalies}\} \cap \{\text{Reported anomalies}\}}{\#\{\text{Real anomalies}\}}$. We perform ten independent trials of injecting contextual anomalies on each dataset and report the means and standard deviations of each of the three evaluation criteria.

\subsection{Anomaly Detection Performance} \label{exp:ADPerformance}

Results on 10 synthetic datasets and 20 real-world datasets are presented in Tables~\ref{tab:SynResults}, \ref{tab:RWResults1}, and \ref{tab:RWResults2}, where the first table concerns synthetic data, and the latter two tables concern real-world data. 


\begin{sidewaystable}[!htbp]
\centering
	\caption{Performance in terms of PRC AUC, ROC AUC, and P@n, on synthetic data, with 10 independent runs of injecting contextual anomalies into each dataset. For each anomaly detector on each dataset, the mean value and standard deviation of each evaluation criterion is presented. The best results obtained on each dataset are highlighted in bold.}
	
	\tiny
\begin{tabular}{ccccccccccc}
		\toprule
		   &  & QCAD & LoPAD & ROCOD & CAD & IForest & LOF & k-NN & SOD & HBOS \\ 
		\midrule 
		\multirow{3}{*}{Syn1}  
		&PRC AUC
		& \textbf{1.00}\textpm0.00 &0.03\textpm0.00 &0.88\textpm0.03 & 0.94\textpm0.15 & 0.32\textpm0.08 & 0.04\textpm0.01 & 0.05\textpm0.01 &0.29\textpm0.03 & 0.18\textpm0.03\\
		&ROC AUC
		&\textbf{1.00}\textpm0.00 &0.49\textpm0.04 & 0.99\textpm0.00 & 0.94\textpm0.16 & 0.95\textpm0.01 & 0.64\textpm0.05 & 0.72\textpm0.03 & 0.97\textpm0.00 &0.90\textpm0.02\\
		&P@n
		&\textbf{0.99}\textpm0.01 &0.04\textpm0.00 & 0.79\textpm0.03 & 0.89\textpm0.31 & 0.37\textpm0.08 & 0.05\textpm0.03 & 0.05\textpm0.02 & 0.20\textpm0.07 &0.22\textpm0.06\\
		\midrule
    	\multirow{3}{*}{Syn2}  
    	&PRC AUC
    	& \textbf{1.00}\textpm0.00 &0.03\textpm0.00 &0.22\textpm0.02 & 0.94\textpm0.03 & 0.15\textpm0.03 & 0.04\textpm0.01 & 0.05\textpm0.01 &0.29\textpm0.02 &0.38\textpm0.08\\
		&ROC AUC
		&\textbf{1.00}\textpm0.00 &0.49\textpm0.02 & 0.95\textpm0.00 & \textbf{1.00}\textpm0.00 & 0.93\textpm0.01 & 0.66\textpm0.03 & 0.71\textpm0.02 & 0.97\textpm0.00 &0.95\textpm0.01\\
		&P@n
		&\textbf{0.98}\textpm0.02 &0.03\textpm0.01 & 0.13\textpm0.03 & 0.94\textpm0.02 & 0.09\textpm0.05 & 0.04\textpm0.02 & 0.04\textpm0.02 & 0.27\textpm0.06 &0.36\textpm0.08\\
		\midrule
    	\multirow{3}{*}{Syn3}  
    	&PRC AUC
    	& \textbf{1.00}\textpm0.00 &0.02\textpm0.00 &0.69\textpm0.04 & 0.96\textpm0.02 & 0.31\textpm0.05 & 0.07\textpm0.01 & 0.06\textpm0.01 &0.34\textpm0.04 &0.26\textpm0.05\\
		&ROC AUC
		&\textbf{1.00}\textpm0.00 &0.47\textpm0.06 & 0.97\textpm0.01 & 0.99\textpm0.01 & 0.93\textpm0.02 & 0.77\textpm0.02 & 0.76\textpm0.02 & 0.97\textpm0.00 &0.92\textpm0.01\\
		&P@n
		&\textbf{0.97}\textpm0.01 &0.02\textpm0.02 & 0.60\textpm0.04 & 0.94\textpm0.02 & 0.34\textpm0.06 & 0.09\textpm0.03 & 0.06\textpm0.02 & 0.34\textpm0.06 &0.29\textpm0.05\\
		\midrule
    	\multirow{3}{*}{Syn4}  
    	&PRC AUC
    	& \textbf{1.00}\textpm0.00 &0.03\textpm0.01 &0.90\textpm0.03 & 0.99\textpm0.01 & 0.38\textpm0.08 & 0.05\textpm0.01 & 0.06\textpm0.01 &0.44\textpm0.04 &0.32\textpm0.04\\
		&ROC AUC
		&\textbf{1.00}\textpm0.00 &0.52\textpm0.02 & \textbf{1.00}\textpm0.00 & \textbf{1.00}\textpm0.00 & 0.95\textpm0.01 & 0.72\textpm0.03 & 0.78\textpm0.03 & 0.99\textpm0.00 &0.93\textpm0.01\\
		&P@n
		&\textbf{1.00}\textpm0.00 &0.03\textpm0.03 & 0.82\textpm0.04 & 0.98\textpm0.02 & 0.43\textpm0.06 & 0.05\textpm0.03 & 0.08\textpm0.03 & 0.49\textpm0.07 &0.37\textpm0.05\\
		\midrule
    	\multirow{3}{*}{Syn5}  
    	&PRC AUC
    	& \textbf{1.00}\textpm0.00 &0.02\textpm0.00 &0.26\textpm0.03 & \textbf{1.00}\textpm0.00 & 0.15\textpm0.02 & 0.04\textpm0.00 & 0.05\textpm0.00 &0.22\textpm0.01 &0.21\textpm0.03\\
		&ROC AUC
		&\textbf{1.00}\textpm0.00 &0.52\textpm0.03 & 0.96\textpm0.00 & \textbf{1.00}\textpm0.00 & 0.93\textpm0.01 & 0.61\textpm0.03 & 0.70\textpm0.02 & 0.96\textpm0.00 &0.94\textpm0.01\\
		&P@n
		&0.99\textpm0.01 &0.02\textpm0.02 & 0.21\textpm0.04 & \textbf{1.00}\textpm0.00 & 0.11\textpm0.06 & 0.04\textpm0.02 & 0.05\textpm0.03 & 0.12\textpm0.03 &0.25\textpm0.05\\
		\midrule
    	\multirow{3}{*}{Syn6}  
    	&PRC AUC
    	& \textbf{0.99}\textpm0.01 &0.03\textpm0.01 &0.97\textpm0.01 & 0.91\textpm0.14 & 0.18\textpm0.04 & 0.03\textpm0.00 & 0.03\textpm0.00 &0.03\textpm0.00 &0.40\textpm0.07\\
		&ROC AUC
		&\textbf{1.00}\textpm0.00 &0.49\textpm0.06 & \textbf{1.00}\textpm0.00 & 0.94\textpm0.16 & 0.88\textpm0.02 & 0.52\textpm0.03 & 0.51\textpm0.04 & 0.53\textpm0.05 &0.93\textpm0.01\\
		&P@n
		&\textbf{0.95}\textpm0.02 &0.03\textpm0.03 & 0.91\textpm0.03 & 0.83\textpm0.29 & 0.26\textpm0.06 & 0.03\textpm0.03 & 0.04\textpm0.03 & 0.02\textpm0.02 &0.42\textpm0.06\\
		\midrule
    	\multirow{3}{*}{Syn7}  
    	&PRC AUC 
    	& \textbf{1.00}\textpm0.00 &0.02\textpm0.00 &0.68\textpm0.07 & 0.98\textpm0.02 & 0.10\textpm0.03 & 0.03\textpm0.00 & 0.03\textpm0.00 &0.03\textpm0.01 &0.28\textpm0.05\\
		&ROC AUC
		&\textbf{1.00}\textpm0.00 &0.48\textpm0.05 & 0.98\textpm0.01 & \textbf{1.00}\textpm0.00 & 0.84\textpm0.03 & 0.49\textpm0.04 & 0.50\textpm0.03 & 0.55\textpm0.03 &0.94\textpm0.01\\
		&P@n
		&\textbf{0.99}\textpm0.01 &0.02\textpm0.02 & 0.64\textpm0.05 & 0.97\textpm0.03 & 0.12\textpm0.05 & 0.02\textpm0.02 & 0.01\textpm0.02 & 0.02\textpm0.03 &0.32\textpm0.04\\
		\midrule
    	\multirow{3}{*}{Syn8}  
    	&PRC AUC 
    	& 0.77\textpm0.04 &0.02\textpm0.00 &0.81\textpm0.04 & \textbf{0.95}\textpm0.03 & 0.17\textpm0.06 & 0.03\textpm0.00 & 0.03\textpm0.00 &0.03\textpm0.01 &0.27\textpm0.06\\
		&ROC AUC
		&0.98\textpm0.01 &0.49\textpm0.04 & 0.98\textpm0.01 & \textbf{0.99}\textpm0.01 & 0.84\textpm0.02 & 0.50\textpm0.03 & 0.52\textpm0.03 & 0.54\textpm0.03 &0.91\textpm0.02\\
		&P@n
		&\textbf{0.91}\textpm0.04 &0.02\textpm0.02 & 0.76\textpm0.04 & 0.90\textpm0.03 & 0.22\textpm0.07 & 0.02\textpm0.02 & 0.02\textpm0.02 & 0.02\textpm0.02 &0.32\textpm0.05\\
		\midrule
    	\multirow{3}{*}{Syn9}  
    	&PRC AUC 
    	& \textbf{0.98}\textpm0.02 &0.02\textpm0.00 &0.92\textpm0.03 & 0.96\textpm0.03 & 0.21\textpm0.03 & 0.03\textpm0.00 & 0.03\textpm0.00 &0.03\textpm0.00 &0.36\textpm0.04\\
		&ROC AUC
		&\textbf{1.00}\textpm0.00 &0.49\textpm0.05 & \textbf{1.00}\textpm0.00 & \textbf{1.00}\textpm0.00 & 0.88\textpm0.02 & 0.50\textpm0.04 & 0.51\textpm0.05 & 0.56\textpm0.04 &0.94\textpm0.02\\
		&P@n
		&\textbf{0.93}\textpm0.03 &0.01\textpm0.01 & 0.87\textpm0.05 & 0.92\textpm0.03 & 0.27\textpm0.07 & 0.02\textpm0.02 & 0.01\textpm0.02 & 0.02\textpm0.02 &0.41\textpm0.05\\
		\midrule
    	\multirow{3}{*}{Syn10}  
    	&PRC AUC 
    	& \textbf{1.00}\textpm0.00 &0.03\textpm0.01 &0.49\textpm0.06 & 0.88\textpm0.04 & 0.11\textpm0.02 & 0.03\textpm0.01 & 0.03\textpm0.01 &0.03\textpm0.01 &0.28\textpm0.05\\
		&ROC AUC
		&\textbf{1.00}\textpm0.00 &0.50\textpm0.04 & 0.98\textpm0.01 & 0.99\textpm0.01 & 0.87\textpm0.02 & 0.51\textpm0.03 & 0.51\textpm0.04 & 0.54\textpm0.05 &0.94\textpm0.01\\
		&P@n
		&\textbf{0.99}\textpm0.01 &0.04\textpm0.03 & 0.45\textpm0.05 & 0.78\textpm0.05 & 0.12\textpm0.04 & 0.02\textpm0.02 & 0.03\textpm0.02 & 0.03\textpm0.01 &0.32\textpm0.04\\
		\bottomrule
		
		\multirow{3}{*}{\textbf{Ranking}\footnotemark[1]}  
		& PRC AUC 
		& 1.2 & 8.6 & 3.0 & 1.9 & 5.1 & 7.4 & 7.1 & 5.6 & 4.6\\
		&ROC AUC 
		& 1.3 & 9.0 & 2.5 & 2.0 & 5.1 & 7.7 & 7.2 & 4.6 & 4.7 \\
		&P@n 
		& 1.1 & 8.0 & 3.2 & 2.0 & 5.0 & 7.3 & 7.3 & 5.9 & 4.3\\
		\bottomrule
	\end{tabular}
	\label{tab:SynResults}

\footnotemark[1]{This is the average ranking of each anomaly detector on $10$  synthetic datasets in terms of PRC AUC, ROC AUC and P@n, respectively.}
\end{sidewaystable}

From Table \ref{tab:SynResults}, we observe that QCAD generally dominates other methods in terms of anomaly detection accuracy according to the average ranks. More specifically, QCAD achieved the best results on 9 out of 10 synthetic datasets in terms of PRC AUC and ROC AUC, and on all datasets in terms of Precision@$n$. On Syn8, QCAD is on par with its best competitor (i.e., CAD) in terms of ROC AUC, whereas QCAD is slightly worse than CAD and ROCOD in terms of PRC AUC. This demonstrates the effectiveness of QCAD in identifying contextual anomalies for different forms and degrees of dependencies between the behavioral features and contextual features. More importantly, the poor performance of LoPAD indicates the importance of distinguishing behavioral features from contextual features. Additionally, other traditional anomaly detectors---including IForest, LOF, $k$-NN, SOD, and HBOS---perform poorly on most datasets because they treat all features equally.

Another important observation is that QCAD is generally superior to other methods in terms of robustness. That is, QCAD attains high PRC AUC, ROC AUC, and Precision@n values with small standard deviations on Syn1, Syn2, Syn3, Syn4 and Syn5, indicating that it is robust to different forms and degrees of dependency relationships, including linearity and non-linearity. In contrast, its strongest contender, CAD, has high standard deviations on Syn1 and Syn6. One possible reason is that CAD gets trapped in a bad local minimum when using expectation-maximization algorithm to learn parameters. Despite being a contextual anomaly detector, ROCOD performs poorly on datasets Syn2 and Syn5 in terms of PRC AUC and Precision@n. From the results on Syn6, Syn7, Syn8, Syn9 and Syn10, it appears that a larger number of contextual features does not substantially affect the performance of QCAD. 
Compared to other contextual anomaly detectors, and specifically CAD, QCAD does not perform particularly better on these synthetic datasets.

\begin{sidewaystable}[!htbp]
\centering
\caption{Performance in terms of PRC AUC, ROC AUC, and P@n, on real-world data, with $5$ or $10$ independent runs of injecting contextual anomalies into each dataset\protect\footnotemark[1]. For each anomaly detector on each dataset, the mean value and standard deviation of each evaluation criterion is presented. The best results obtained on each dataset are highlighted in bold.}
\tiny
\begin{tabular}{cccc cccc ccc}
		\toprule
		   &  & QCAD & LoPAD & ROCOD & CAD\footnotemark[2] & IForest & LOF & k-NN & SOD & HBOS \\ 
		\midrule 
		\multirow{3}{*}{Abalone}  & PRC AUC 
		& \textbf{0.96}\textpm0.01 & 0.02\textpm0.00 & 0.55\textpm0.04 & 0.27 &0.39\textpm0.05 &0.92\textpm0.01 &0.94\textpm0.00 &0.75\textpm0.02 &0.19\textpm0.03\\
		&ROC AUC 
		&\textbf{1.00}\textpm0.00 & 0.36\textpm0.05 & 0.98\textpm0.00 & 0.93 &0.98\textpm0.00 &\textbf{1.00}\textpm0.00 & \textbf{1.00}\textpm0.00 &0.96\textpm0.01 &0.91\textpm0.01\\
		&P@n 
		&0.90\textpm0.02 & 0.02\textpm0.00 & 0.58\textpm0.01 & 0.40 &0.41\textpm0.07 &0.92\textpm0.02 &\textbf{0.93}\textpm0.00 &0.74\textpm0.02 &0.19\textpm0.06\\
		\midrule 
		\multirow{3}{*}{Airfoil}  & PRC AUC
		& \textbf{0.69}\textpm0.05 & 0.05\textpm0.01 & 0.41\textpm0.05 & 0.46\textpm0.04 &0.10\textpm0.02 &0.05\textpm0.01 &0.05\textpm0.01 &0.14\textpm0.02 &0.09\textpm0.02\\
		&ROC AUC
		&\textbf{0.91}\textpm0.03 & 0.51\textpm0.03 & 0.79\textpm0.02 & 0.76\textpm0.03 &0.72\textpm0.03 &0.53\textpm0.04 & 0.52\textpm0.02 &0.78\textpm0.02 &0.66\textpm0.02\\
		&P@n 
		&\textbf{0.66}\textpm0.05 & 0.04\textpm0.03 & 0.40\textpm0.04 & 0.45\textpm0.02 &0.13\textpm0.04 &0.05\textpm0.03 &0.06\textpm0.03 &0.14\textpm0.03 &0.12\textpm0.02\\
		\midrule 
		\multirow{3}{*}{BodyFat}  & PRC AUC
		&0.60\textpm0.11 & 0.08\textpm0.02 &0.54\textpm0.00 & \textbf{0.92}\textpm0.08 &0.16\textpm0.03 &0.09\textpm0.02  &0.09\textpm0.02 &0.10\textpm0.04 &0.18\textpm0.04\\
		&ROC AUC
		&0.91\textpm0.05 & 0.48\textpm0.04 & 0.50\textpm0.00 & \textbf{0.99}\textpm0.00 &0.73\textpm0.04 &0.51\textpm0.06 &0.52\textpm0.05 & 0.53\textpm0.08&0.76\textpm0.04\\
		&P@n 
		&0.58\textpm0.12 & 0.06\textpm0.06 & 0.08\textpm0.03 & \textbf{0.88}\textpm0.05 &0.16\textpm0.06 &0.08\textpm0.08 &0.10\textpm0.06 &0.10\textpm0.04&0.23\textpm0.07\\
		\midrule 
		\multirow{3}{*}{Boston}  & PRC AUC
		& \textbf{0.72}\textpm0.05 & 0.08\textpm0.01 & 0.30\textpm0.06 & 0.33\textpm0.13 &0.11\textpm0.02 &0.08\textpm0.01 &0.08\textpm0.02 &0.08\textpm0.01 &0.13\textpm0.03\\
		&ROC AUC
		&\textbf{0.92}\textpm0.03 & 0.48\textpm0.03 & 0.83\textpm0.04 & 0.70\textpm0.11 &0.60\textpm0.04 &0.48\textpm0.04 & 0.49\textpm0.04 &0.51\textpm0.04 &0.66\textpm0.04\\
		&P@n 
		&\textbf{0.68}\textpm0.04 & 0.07\textpm0.02 & 0.37\textpm0.05 & 0.25\textpm0.10 &0.13\textpm0.03 &0.07\textpm0.04 &0.07\textpm0.04 &0.06\textpm0.03 &0.15\textpm0.06\\
		\midrule 
		\multirow{3}{*}{Concrete}  & PRC AUC
		&\textbf{0.63}\textpm0.06 & 0.09\textpm0.01 &0.33\textpm0.06 & 0.33\textpm0.07 &0.08\textpm0.02 &0.06\textpm0.01 &0.06\textpm0.01 &0.06\textpm0.01 &0.25\textpm0.06\\
		&ROC AUC
		&\textbf{0.93}\textpm0.03 & 0.62\textpm0.03 & 0.77\textpm0.04 & 0.70\textpm0.04 &0.61\textpm0.03 &0.49\textpm0.06 & 0.50\textpm0.05 &0.50\textpm0.03 &0.72\textpm0.05\\
		&P@n 
		&\textbf{0.62}\textpm0.05 & 0.1\textpm0.03 & 0.33\textpm0.04 & 0.33\textpm0.06 &0.08\textpm0.03 &0.06\textpm0.03 &0.06\textpm0.02 &0.05\textpm0.02 &0.29\textpm0.04\\
		\midrule
		\multirow{3}{*}{El Nino}  & PRC AUC
		&\textbf{0.98}\textpm0.01 & 0.11\textpm0.01 &0.40\textpm0.02 & 0.57 &0.53\textpm0.08 &0.01\textpm0.00 &0.01\textpm0.00 &0.04\textpm0.00 &0.77\textpm0.03\\
		&ROC AUC
		&\textbf{1.00}\textpm0.00 & 0.91\textpm0.01 & 0.98\textpm0.01 & 0.91 &0.97\textpm0.01 &0.50\textpm0.02 & 0.50\textpm0.03 &0.90\textpm0.00 &0.99\textpm0.00\\
		&P@n 
		&\textbf{0.93}\textpm0.02 & 0.13\textpm0.03 & 0.42\textpm0.03 & 0.55 &0.50\textpm0.06 &0.01\textpm0.01 &0.01\textpm0.01 &0.01\textpm0.01 &0.70\textpm0.02\\		
		\midrule 
		\multirow{3}{*}{Energy}  & PRC AUC
		& \textbf{0.92}\textpm0.02 & 0.05\textpm0.01 & 0.42\textpm0.06 & 0.32\textpm0.05 &0.35\textpm0.07 &0.10\textpm0.03 &0.37\textpm0.02 &0.89\textpm0.04 &0.31\textpm0.05\\
		&ROC AUC
		& \textbf{0.99}\textpm0.00 & 0.40\textpm0.05 & 0.82\textpm0.03 & 0.62\textpm0.04 &0.88\textpm0.03 &0.55\textpm0.04 & 0.95\textpm0.00 &0.98\textpm0.00 &0.79\textpm0.02\\
		&P@n 
		& \textbf{0.81}\textpm0.04 & 0.02\textpm0.01 & 0.42\textpm0.04 & 0.32\textpm0.06 &0.43\textpm0.07 &0.11\textpm0.04 &0.30\textpm0.03 &0.79\textpm0.06 &0.35\textpm0.06\\
		\midrule 
		\multirow{3}{*}{FishWeight}  & PRC AUC
		&\textbf{0.81}\textpm0.05 & 0.13\textpm0.06 & 0.46\textpm0.10 & 0.41\textpm0.12 & 0.23\textpm0.07 &0.12\textpm0.04 &0.12\textpm0.04 &0.11\textpm0.03 &0.21\textpm0.06\\
		&ROC AUC
		&\textbf{0.96}\textpm0.02 & 0.52\textpm0.05 & 0.82\textpm0.07 & 0.69\textpm0.11 & 0.77\textpm0.06 &0.51\textpm0.10 & 0.52\textpm0.09 &0.56\textpm0.08 &0.69\textpm0.06\\
		&P@n 
		&\textbf{0.71}\textpm0.10 & 0.13\textpm0.09 & 0.44\textpm0.08 & 0.45\textpm0.12 &0.31\textpm0.11 &0.14\textpm0.11 &0.13\textpm0.10 &0.09\textpm0.06 &0.30\textpm0.13\\
		\midrule 
		\multirow{3}{*}{ForestFires} 
		&PRC AUC
		& 0.99\textpm0.00 &0.10\textpm0.01 &0.48\textpm0.05 & 0.29\textpm0.07 & 0.97\textpm0.02 & 0.17\textpm0.02 & 0.18\textpm0.02 &0.58\textpm0.04 & \textbf{1.00}\textpm0.00\\
		&ROC AUC
		& \textbf{1.00}\textpm0.00 &0.50\textpm0.04 & 0.88\textpm0.02 & 0.85\textpm0.05 & \textbf{1.00}\textpm0.00 & 0.71\textpm0.04 & 0.74\textpm0.01 & 0.96\textpm0.01) & \textbf{1.00}\textpm0.00\\
		&P@n 
		& 0.96\textpm0.02 &0.09\textpm0.04 & 0.41\textpm0.04 & 0.28\textpm0.12 & 0.94\textpm0.02 & 0.16\textpm0.05 &0.12\textpm0.04 & 0.60\textpm0.04 & \textbf{0.99}\textpm0.01\\
		\midrule 
		\multirow{3}{*}{GasEmission}  & PRC AUC
		& \textbf{0.94}\textpm0.02 & 0.01\textpm0.00 & 0.86\textpm0.02 & 0.63 &0.10\textpm0.01 &0.01\textpm0.00 &0.01\textpm0.00 &0.03\textpm0.00 &0.27\textpm0.07\\
		&ROC AUC
		& \textbf{1.00}\textpm0.00 & 0.47\textpm0.01 & \textbf{1.00}\textpm0.00 & 0.99 &0.93\textpm0.01 &0.52\textpm0.04 & 0.52\textpm0.01 &0.78\textpm0.01) &0.97
		\textpm0.00\\
		&P@n 
		& \textbf{0.89}\textpm0.02 & 0.00\textpm0.00 & 0.83\textpm0.02 & 0.62 &0.05\textpm0.02 &0.02\textpm0.01 &0.02\textpm0.02 &0.02\textpm0.02 &0.24\textpm0.05\\
		\midrule 
		\multirow{3}{*}{HeartFailure}  & PRC AUC
		& 0.90\textpm0.03 & 0.05\textpm0.00 &0.53\textpm0.04 & 0.82\textpm0.06 &0.86\textpm0.04 &0.13\textpm0.02 &0.18\textpm0.04 &0.38\textpm0.07 & \textbf{0.97}\textpm0.01\\
		&ROC AUC
		& 0.98\textpm0.01 & 0.10\textpm0.03 & 0.52\textpm0.05 & 0.98\textpm0.01 &0.98\textpm0.01 &0.57\textpm0.06 & 0.68\textpm0.05 &0.88\textpm0.02 & \textbf{1.00}\textpm0.00\\
		&P@n 
		& 0.83\textpm0.05 & 0.00\textpm0.00 & 0.16\textpm0.14 & 0.78\textpm0.05 &0.80\textpm0.05 &0.12\textpm0.06 &0.20\textpm0.07 &0.34\textpm0.07 & \textbf{0.92}\textpm0.03\\
		\midrule 
		\multirow{3}{*}{Hepatitis}  & PRC AUC
		& 0.98\textpm0.01 & 0.05\textpm0.01 &0.55\textpm0.03 & 0.79\textpm0.07 &0.92\textpm0.03 &0.17\textpm0.02 &0.14\textpm0.02 &0.35\textpm0.01 & \textbf{0.99}\textpm0.01\\
		&ROC AUC
		& \textbf{1.00}\textpm0.00 & 0.49\textpm0.06 & 0.98\textpm0.00 & 0.99\textpm0.00 &\textbf{1.00}\textpm0.00 &0.87\textpm0.02 & 0.84\textpm0.01 &0.96\textpm0.00 & \textbf{1.00}\textpm0.00\\
		&P@n 
		& 0.92\textpm0.02 & 0.05\textpm0.02 & 0.65\textpm0.04 & 0.78\textpm0.06 &0.89\textpm0.02 &0.06\textpm0.04 &0.07\textpm0.04 &0.33\textpm0.03 & \textbf{0.96}\textpm0.01\\
		\midrule
\end{tabular}
\label{tab:RWResults1}
\end{sidewaystable}

\begin{sidewaystable}[!htbp]
\centering
\caption{(Table \ref{tab:RWResults1} Continued) Performance in terms of PRC AUC, ROC AUC, and P@n, on real-world data, with $5$ or $10$ independent runs of injecting contextual anomalies into each dataset\protect\footnotemark[1]. For each anomaly detector on each dataset, the mean value and standard deviation of each evaluation criterion is presented. The best results obtained on each dataset are highlighted in bold.}
\tiny
\begin{tabular}{cccc cccc ccc}
		\toprule
		   &  & QCAD & LoPAD & ROCOD & CAD\footnotemark[2] & IForest & LOF & k-NN & SOD & HBOS \\ 
		\midrule 
		\multirow{3}{*}{IndianLiver}  & PRC AUC
		& 0.90\textpm0.05 & 0.05\textpm0.00 &0.48\textpm0.06 & 0.84\textpm0.12 &0.93\textpm0.02 &0.11\textpm0.02 &0.17\textpm0.04 &0.44\textpm0.06 & \textbf{0.99}\textpm0.01\\
		&ROC AUC
		& \textbf{1.00}\textpm0.00 & 0.48\textpm0.05 & 0.97\textpm0.01 & 0.94\textpm0.16 & \textbf{1.00}\textpm0.00 &0.72\textpm0.05 & 0.81\textpm0.04 &0.96\textpm0.01 & \textbf{1.00}\textpm0.00\\
		&P@n 
		& 0.85\textpm0.03 & 0.05\textpm0.02 & 0.47\textpm0.07 & 0.74\textpm0.25 &0.90\textpm0.02 &0.07\textpm0.03 &0.20\textpm0.10 &0.40\textpm0.07 & \textbf{0.95}\textpm0.02\\
		\midrule 
		\multirow{3}{*}{Maintenance}  & PRC AUC
		& \textbf{0.91}\textpm0.01 & 0.01\textpm0.00 & 0.50\textpm0.00 & 0.50 &0.02\textpm0.01 &0.01\textpm0.00 &0.01\textpm0.00 &0.02\textpm0.00 &0.19\textpm0.08\\
		&ROC AUC
		& \textbf{1.00}\textpm0.00 & 0.48\textpm0.05 & 0.50\textpm0.00 & 0.50 &0.70\textpm0.02 &0.49\textpm0.01 & 0.55\textpm0.02 &0.79\textpm0.01 &0.67\textpm0.04\\
		&P@n 
		& \textbf{0.84}\textpm0.02 & 0.01\textpm0.02 & 0.04\textpm0.01 & 0.00 &0.03\textpm0.02 &0.00\textpm0.00 &0.01\textpm0.00 &0.01\textpm0.01 &0.32\textpm0.09\\
		\midrule 
		\multirow{3}{*}{Parkinson}  & PRC AUC
		& \textbf{0.81}\textpm0.01 & 0.02\textpm0.01 & 0.58\textpm0.06 & 0.68 & 0.02\textpm0.00 &0.02\textpm0.00 &0.02\textpm0.00 &0.04\textpm0.00 &0.02\textpm0.00\\
		&ROC AUC
		& \textbf{0.99}\textpm0.00 & 0.43\textpm0.03 & 0.94\textpm0.02 & 0.90 & 0.56\textpm0.03 &0.48\textpm0.03 & 0.48\textpm0.02 &0.75\textpm0.02 &0.59\textpm0.03\\
		&P@n 
		&\textbf{0.85}\textpm0.01 & 0.02\textpm0.01 & 0.54\textpm0.05 & 0.62 & 0.01\textpm0.01 &0.04\textpm0.01 &0.06\textpm0.01 &0.02\textpm0.01 &0.01\textpm0.01\\
		\midrule
		\multirow{3}{*}{PowerPlant}  & PRC AUC
		&0.56\textpm0.03 & 0.02\textpm0.00 &\textbf{0.65}\textpm0.05 & 0.32 &0.04\textpm0.00 &0.01\textpm0.00 &0.01\textpm0.00 &0.01\textpm0.00 &0.22\textpm0.03\\
		&ROC AUC
		&\textbf{0.98}\textpm0.01 & 0.45\textpm0.01 & 0.97\textpm0.01 & 0.64 &0.78\textpm0.03 &0.50\textpm0.03 & 0.49\textpm0.02 &0.63\textpm0.03 &0.71\textpm0.04\\
		&P@n 
		&0.58\textpm0.05 & 0.01\textpm0.01 & \textbf{0.68}\textpm0.01 & 0.31 &0.07\textpm0.02 &0.01\textpm0.01 &0.02\textpm0.02 &0.01\textpm0.01 & 0.27\textpm0.04\\
		\midrule 
		\multirow{3}{*}{QSRanking}  & PRC AUC
		& \textbf{0.79}\textpm0.04 & 0.06\textpm0.00 & 0.57\textpm0.08 & 0.24\textpm0.07 &0.09\textpm0.03 &0.09\textpm0.03 &0.09\textpm0.02 &0.09\textpm0.01 &0.47\textpm0.08\\
		&ROC AUC
		& \textbf{0.96}\textpm0.01 & 0.37\textpm0.02 & 0.89\textpm0.03 & 0.74\textpm0.04 &0.48\textpm0.07 &0.48\textpm0.07 & 0.48\textpm0.04 &0.51\textpm0.04 &0.87\textpm0.03\\
		&P@n 
		& \textbf{0.72}\textpm0.05 & 0.02\textpm0.02 & 0.53\textpm0.08 & 0.26\textpm0.09 &0.08\textpm0.05 &0.08\textpm0.04 &0.08\textpm0.05 &0.08\textpm0.03 &0.48\textpm0.07\\
		\midrule 
		\multirow{3}{*}{SynMachine}  & PRC AUC
		& \textbf{0.96}\textpm0.03 & 0.34\textpm0.04 & 0.79\textpm0.06 & 0.42\textpm0.07 &0.34\textpm0.05 &0.80\textpm0.04 &0.91\textpm0.00 &0.71\textpm0.06 &0.26\textpm0.03\\
		&ROC AUC
		& 0.98\textpm0.01 & 0.78\textpm0.03 & 0.89\textpm0.03 & 0.65\textpm0.07 &0.84\textpm0.03 &0.92\textpm0.03 & \textbf{0.99}\textpm0.01 &0.85\textpm0.03 &0.68\textpm0.02\\
		&P@n 
		& \textbf{0.94}\textpm0.03 & 0.39\textpm0.05 & 0.73\textpm0.07 & 0.34\textpm0.11 &0.39\textpm0.05 &0.72\textpm0.04 &0.78\textpm0.03 &0.65\textpm0.04 &0.27\textpm0.05\\
		\midrule 
		\multirow{3}{*}{Toxicity}  & PRC AUC
		&\textbf{0.46}\textpm0.09 & 0.10\textpm0.01 &\textbf{0.46}\textpm0.06 & 0.50\textpm0.06 &0.10\textpm0.01 &0.08\textpm0.02 &0.07\textpm0.01 &0.12\textpm0.02 &0.10\textpm0.01\\
		&ROC AUC
		&\textbf{0.88}\textpm0.03 & 0.57\textpm0.03 & 0.84\textpm0.03 & 0.56\textpm0.13 &0.71\textpm0.04 &0.60\textpm0.04 & 0.60\textpm0.03 &0.73\textpm0.06 &0.69\textpm0.03\\
		&P@n 
		&\textbf{0.49}\textpm0.07 & 0.18\textpm0.01 & 0.47\textpm0.06 & 0.12\textpm0.15 &0.09\textpm0.04 &0.06\textpm0.03 &0.05\textpm0.01 &0.11\textpm0.04 &0.09\textpm0.04\\
		\midrule 
		\multirow{3}{*}{Yacht}  & PRC AUC
		&\textbf{0.90}\textpm0.04 & 0.53\textpm0.02 &0.29\textpm0.05 & 0.55\textpm0.00 &0.28\textpm0.08 &0.20\textpm0.04 &0.25\textpm0.03 &0.24\textpm0.05 &0.32\textpm0.08\\
		&ROC AUC
		&\textbf{0.98}\textpm0.02 & 0.96\textpm0.00 & 0.78\textpm0.03 & 0.50\textpm0.00 &0.78\textpm0.06 &0.72\textpm0.05 & 0.83\textpm0.02 &0.59\textpm0.07 &0.76\textpm0.05\\
		&P@n 
		&\textbf{0.80}\textpm0.06 & 0.58\textpm0.06 & 0.30\textpm0.06 & 0.08\textpm0.04 &0.32\textpm0.08 &0.19\textpm0.05 &0.24\textpm0.04 &0.25\textpm0.05 &0.30\textpm0.07\\
		\bottomrule
		\multirow{3}{*}{\textbf{Ranking} \footnotemark[3]}
		& PRC AUC
          & 1.40 & 7.45 & 3.30 & 3.45 & 4.90 & 7.20 & 6.75 & 5.75 & 4.25\\
		&ROC AUC
		& 1.40 & 7.95 & 3.70 & 5.00 & 4.15 & 7.00 & 6.05 & 5.00 &4.10 \\
		&P@n 
		& 1.45 & 7.35 & 3.65 & 4.00 & 4.60 & 6.95 &6.25 & 6.05 & 4.10\\
		\bottomrule
\end{tabular}

\footnotemark[1]{Due to computation time limitations, we perform 5 independent trials for datasets with a sample size greater than 2000. For smaller datasets, we conduct 10 independent trials.}

\footnotemark[2]{We only conduct an independent experiment using CAD on Maintenance, Gas Emission, Parkinson Telemonitoring, Power Plant, and Elnino. It would take more than 1 week for CAD to complete 5 independent trials on each of these individual datasets.}

\footnotemark[3]{This is the average ranking of each anomaly detector on $20$  real-world datasets in terms of PRC AUC, ROC AUC, or P@n, respectively. 
}
\label{tab:RWResults2}
\end{sidewaystable}

The results on real-world datasets presented in Tables~\ref{tab:RWResults1} and~\ref{tab:RWResults2} show that QCAD performs best overall when compared to its contenders, as witnessed by its average ranking results.
Concretely, QCAD outperforms the other methods on 13 out of 20 real-world datasets in terms of PRC AUC and ROC AUC, and on 11 out of 20 datasets in terms of P@n. On most of the remaining datasets, QCAD is on par with its strongest competitors. For instance, HBOS achieves the best performance on Forest Fires, Heart Failures, Hepatitis, and Indian Liver Patient; QCAD's performance on these datasets is comparable. The number of contextual features in these datasets is often substantially less than the number of behavioral features, reducing the contextual anomaly detection problem to a traditional anomaly detection problem. QCAD is only slightly worse than ROCOD on Power Plant and Toxicity in terms of PRC AUC. Note that CAD produces surprisingly good results on the Bodyfat dataset, and surpasses other methods including QCAD. One possible explanation is that the contexts and behaviors are both composed of well-separated Gaussian components, and that the association between contextual and behavioral components is strong.

\begin{figure}[!htbp]
\centering
\includegraphics[width=12cm]{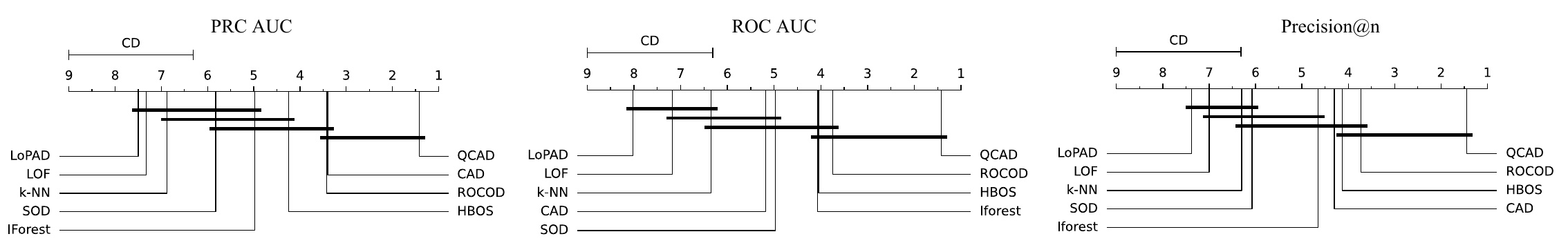}
\caption{Critical difference diagram showing statistical difference comparisons between QCAD and its contenders in terms of PRC AUC, ROC AUC and Precision@n. To achieve this, we use Friedman tests \citep{friedman1937use} followed by Nemenyi post hoc analysis \citep{nemenyi1963distribution} with a significance level of $0.05$. The post-hoc Nemenyi test indicates there are no significant differences within QCAD, CAD and ROCOD in terms of PRC AUC; there are no significant differences within QCAD, ROCOD, HBOS and IForest in terms of ROC AUC; there are no significant differences within QCAD, ROCOD, HBOS and CAD in terms of Precision@n. However, one can see that QCAD consistently outperforms its contenders by a large margin in three metrics.
}\label{fig:StatCom}
\end{figure}

We also observe that QCAD performs well on datasets with varying sample size, dimensionality, and rate of injected anomalies. Specifically, QCAD achieved a high ROC AUC ($\ge 0.85$) on all datasets. This is much better than the ROC AUC values close to $0.50$ obtained by ROCOD, CAD, HBOS, and IForest on some datasets, implying they are random guessing. In addition, QCAD attained high PRC AUC ($\ge 0.8$) and Precision@n values ($\ge 0.7$) on most datasets. The lowest PRC AUC and Precision@n value are is 0.46 and 0.49, respectively, both obtained on the Toxicity dataset. Possible reasons for QCAD's moderate performance on Airfoil, BodyFat, Power plant, and Toxicity are: the dependency relationship between behavioral features and contextual features is not strong, or the dependency relationship is too complex for the Quantile Regression Forests to capture. ROCOD, COD, HBOS, and IForest, however, obtain PRC AUC values less than 0.70 and Precision@n values less than 0.60 on most datasets. Moreover, HBOS and IForest sometimes attain PRC AUC and Precision@n lower than 0.10, which is extremely poor. This implies that traditional anomaly detection methods are not suitable for identifying contextual anomalies, as they treat all features equally. Furthermore, the relatively poor performance of ROCOD and CAD---compared to our method---demonstrates the importance of modelling more properties of the conditional distribution than just the mean.

\subsection{Runtime Analysis} \label{subsec:runtime}
\begin{figure}[!htbp]
\centering
\includegraphics[width=12cm]{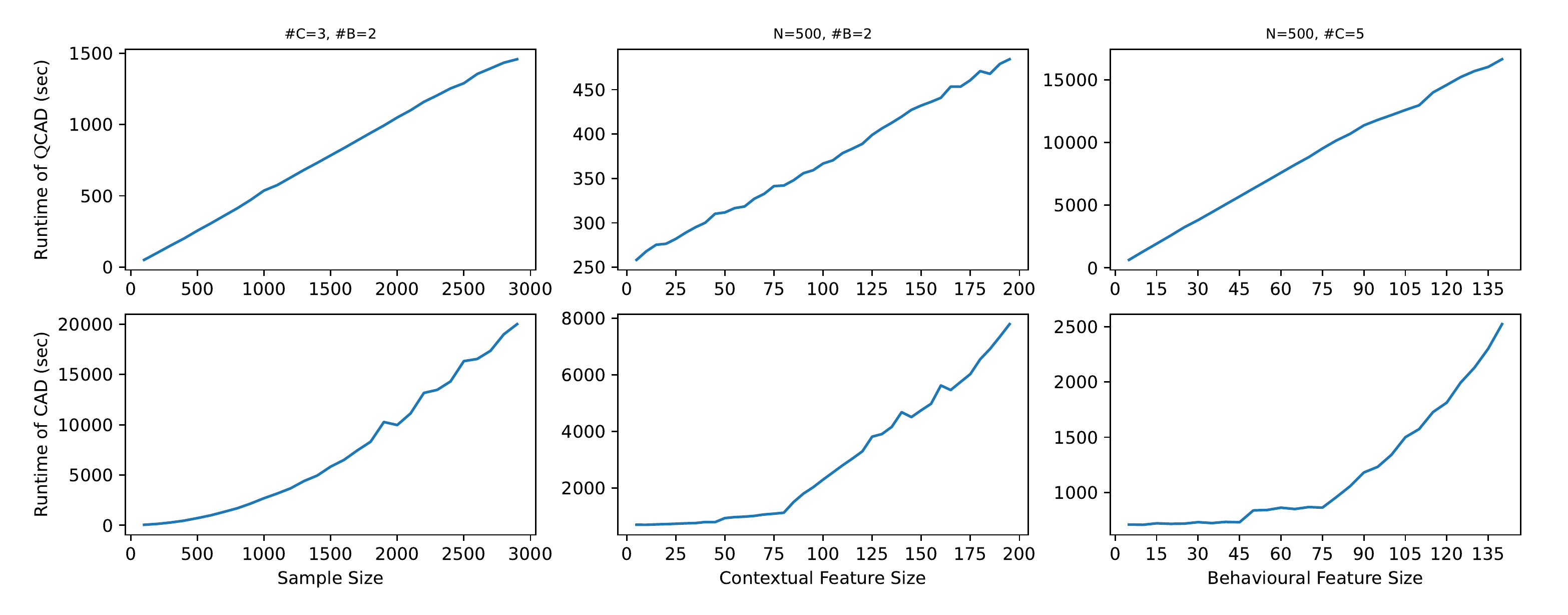}
\caption{Runtime analysis by varying the sample size, the number of contextual features ($\#\textbf{C}$), and the number of behavioral features ($\#\textbf{B}$). The results are obtained with 5 independent trials. The data was synthesized using scheme S1. Note the varying y-axis scales.}\label{fig:runtime}
\end{figure}

To investigate the scalability and efficiency of QCAD, we perform a runtime analysis by varying the sample size, the number of contextual features, and the number of behavioral features. As shown in Figure~\ref{fig:runtime}, we can empirically observe that the runtime of QCAD scales linearly with respect to the sample size, the number of contextual features, and the number of behavioral features on small and medium datasets. In contrast, CAD is computationally prohibitively expensive even on small datasets. To further understand the complexity of QCAD, we perform a time complexity analysis in Appendix~\ref{Appendix:Complexity}. The theoretical analysis is in line with our empirical results and analysis.

\subsection{Using QCAD to Find Promising Football Players} 
\label{subsec:application}

In this section, we illustrate the capability of QCAD on identifying potentially meaningful contextual anomalies and providing intuitive and understandable explanations for reported anomalies when applied to a real-world problem.

\begin{figure}[!htbp]
\centering
\includegraphics[width=8cm]{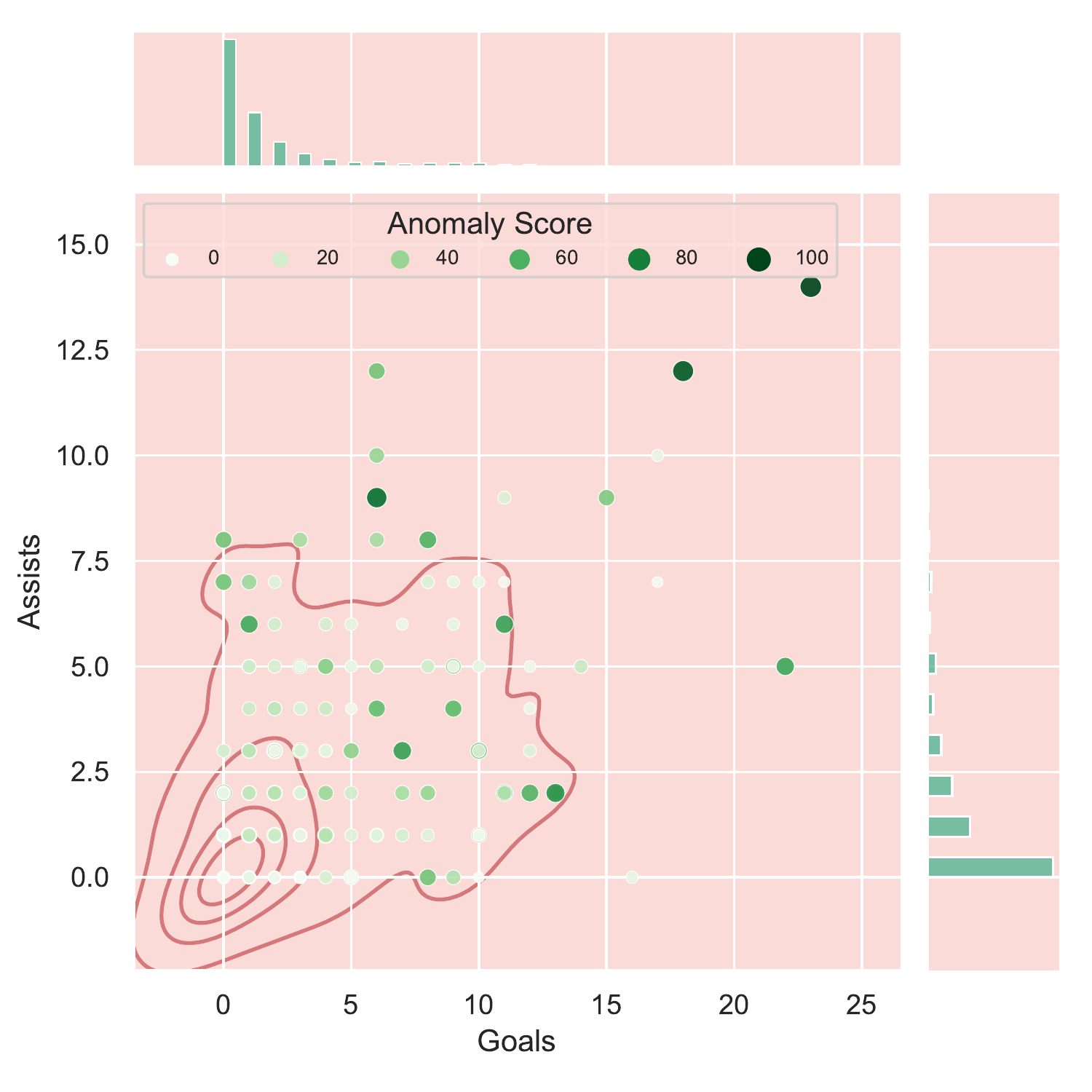}
\caption{Distribution of all players in the behavioral space $Assists \times Goals$, and their corresponding anomaly scores given by QCAD. The red curves represent the estimated densities of the joint distribution at different levels, the histograms indicate the marginal distributions. A larger and darker green dot represents a larger anomaly score as given by QCAD. }\label{fig:football1}
\end{figure}

As use case, we consider the problem of finding exceptional players in the English Premier League. The dataset\footnote{https://www.kaggle.com/rajatrc1705/english-premier-league202021} describes the background information and performance statistics of 532 football players from 2020 to 2021. We use the variables \textit{Position1,Position2} (positions for which the player plays), \textit{Age} (age of the player), \textit{Matches} (number of matches played), \textit{Starts} (number of matches that the player was in the starting lineup), and  \textit{Mins} (number of minutes the player played overall) as contextual features. Two performance statistics, \textit{Goals} (number of goals scored by the player) and \textit{Assists} (number of assists given by player), are regarded as behavioral features.

Figure \ref{fig:football1} depicts the distribution of all data objects in behavioral space $Goals\times Assists$. Without considering contextual information, traditional anomaly detectors such as HBOS, LOF, IForest will treat objects in dense areas (i.e., objects residing inside the red curves) as normal objects. In contrast, our contextual anomaly detector QCAD can detect anomalies in dense areas by considering contextual information (i.e., green objects residing inside the red curves). Concretely, Figure~\ref{fig:football1}  also presents the anomaly scores given by QCAD (with default settings). For example, player Matheus Pereira---who achieved 11 goals and 6 assists---is reported as an `anomaly' (i.e., as having a relatively high anomaly score) by QCAD but as `normal' by traditional anomaly detectors. Conversely, QCAD considers Patrick Bamford, a player with 17 goals and 7 assists (in a sparse area), to be normal, while traditional anomaly detectors consider him to be an anomaly.

\begin{figure}[!htbp]
\centering
\includegraphics[width=12cm]{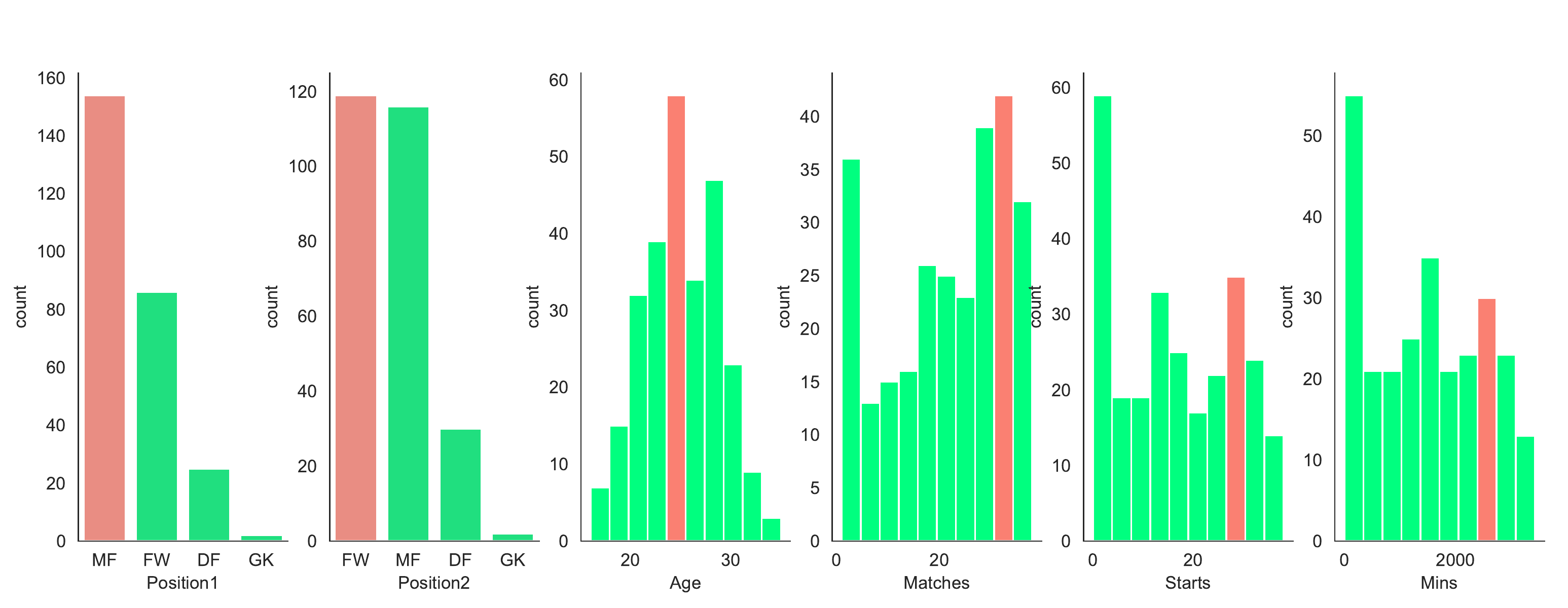}
\caption{Explanations for the identified anomalous player Matheus Pereira based on his reference group. Shown are the values for Matheus Pereira (red bins) relative to the distribution of his contextual neighbours for each contextual feature.}\label{fig:football2}
\end{figure}

\begin{figure}[!htbp]
\centering
\includegraphics[width=12cm]{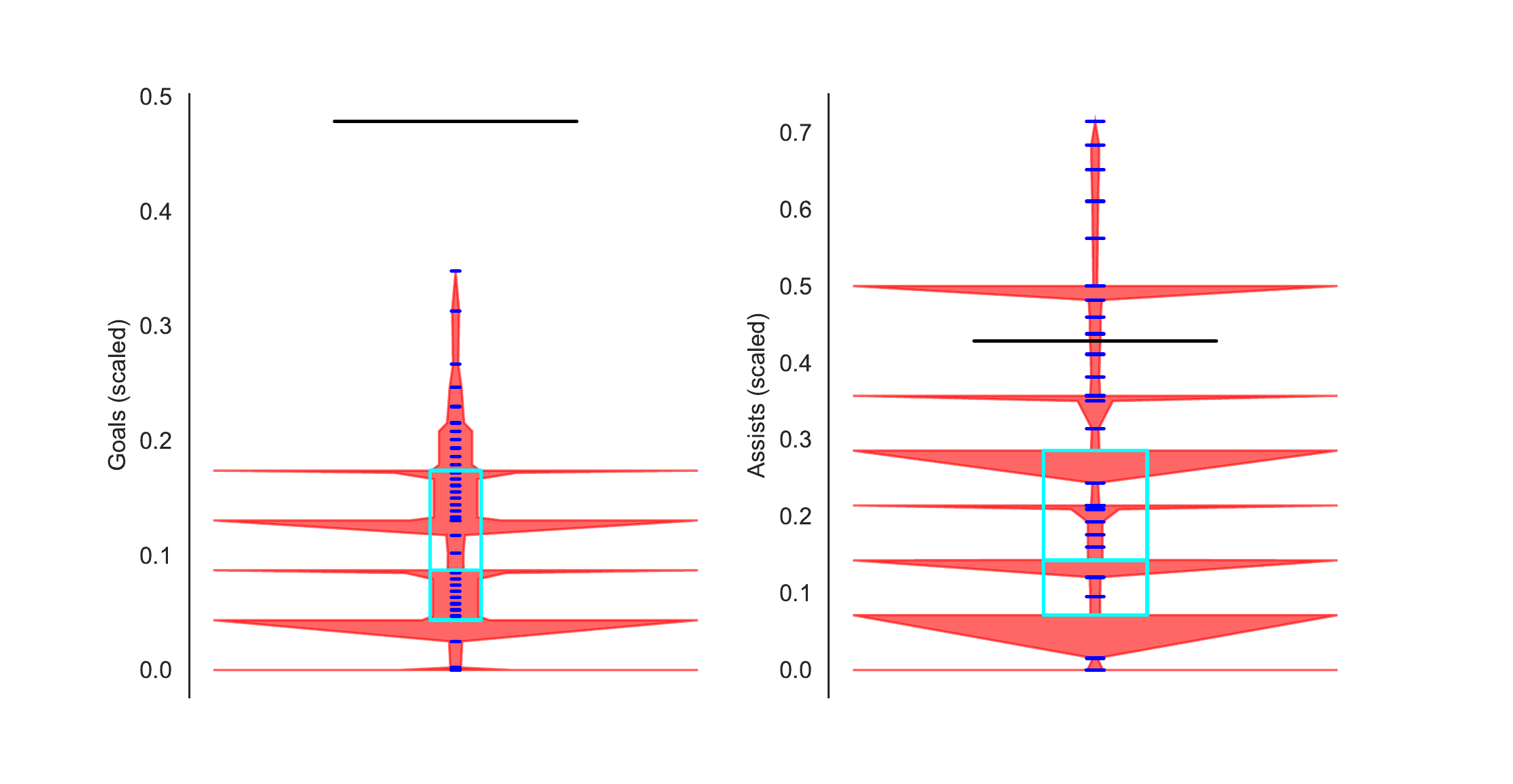}
\caption{Explaining why Matheus Pereira is considered an anomaly: exceptionally many goals compared to contextually similar players, and relatively many assists. The beanplot shows the estimated conditional distribution of each behavioral feature, with a wider red area representing a higher probability of occurrence. The horizontal black lines represent  the values reported for the player being investigated.}\label{fig:football3}
\end{figure}

In addition to giving an anomaly score for each data object, QCAD can also provide an explanation. For instance, for Matheus Pereira ($Position1$ = "MF", $Position2$ = "FW", $Age$=24, $Matches$=33, $Starts$=30, $Mins$=2577), Figure~\ref{fig:football2} shows the distribution of the values of each contextual feature for all players in his contextual neighbourhood (i.e., similar players). It can be seen that most of these players are Midfielder (MF) and/or Forward(FW), aged from 22 to 28, playing matches more than 25 times, etc. The anomaly score is 65.8, which puts this player in the top-$10$ of most anomalous players. This score can be decomposed according to the behavioral features to give the behavioral feature(s) that contribute the most to the obtained anomaly score; \emph{Goals}, in this case. Moreover, Figure~\ref{fig:football3} shows how the beanplot-like visualisation can help to give insight in how the behavioral feature values deviate from those in the player's contextual neighbourhood. From these plots we can see that Matheus Pereira has been exceptionally good at scoring goals and slightly better than expected with regard to giving assists.

These results can be interpreted as: Matheus Pereira performed surprisingly well in terms of goals and slightly better than expected in terms of assists when compared to other midfielders and/or forwards aged 22-28 who played more than 25 games and played more than 2000 minutes. Although we are not football experts, we expect the automated detection and interpretable score explanation of such `anomalies' to be meaningful and possibly useful to football coaches and scouts. 

%

\section{Conclusions}
\label{sec:conclusion}

In this paper,  we for the first time explicitly establish a connection between dependency-based traditional anomaly detection methods and contextual anomaly detection methods. On this basis, we propose a novel approach to contextual anomaly detection and explanation. Specifically, we use Quantile Regression Forests to develop a accurate and interpretable anomaly detection method, QCAD, that explores dependencies between features. QCAD can handle tabular datasets with mixed contextual features and numerical behavioral features. Extensive experiment results on various synthetic and real-world datasets demonstrate that QCAD outperforms state-of-the-art anomaly detection methods in identifying contextual anomalies in terms of accuracy and interpretability. 

From the case study on football player data, we conclude that QCAD can detect potentially meaningful and useful contextual anomalies that are directly interpretable by a domain expert. The beanplot-based visualisations help to explain why a certain object is (not) considered an anomaly within its context. This is important because anomaly detection is an unsupervised problem and the explanations can help analysts and domain experts to verify the results. It also opens up opportunities for human-guided anomaly detection, where feedback from the analyst can be used to guide the analysis.

In the future, given that QCAD can only handle static features, we plan to extend QCAD to streaming settings.
 
\section*{Statement and Declaration}
\backmatter


\bmhead{Ethical approval}The human and animal data involved in this study are publicly available, and thus the need for approval was waived.

\bmhead{Funding}
This work is supported by Project 4 of the Digital Twin research programme, a TTW Perspectief programme with project number P18-03 that is primarily financed by the Dutch Research Council (NWO). All opinions, findings, conclusions and recommendations in this paper are those of the authors and do not necessarily reflect the views of the funding agencies.

\bmhead{Conflict of interest} The author(s) declared no potential conflicts of interest with
respect to the research, authorship and/or publication of this
article.

\bmhead{Availability of data and materials}
For reproducibility, all code and datasets are provided online via the following link: \href{https://github.com/ZhongLIFR/QCAD}{https://github.com/ZhongLIFR/QCAD}.

\bmhead{Authorship Contribution}

\textit{Zhong Li}: Conceptualization, Methodology, Validation, Investigation, Writing, Visualisation, Project Administration. \textit{Matthijs van Leeuwen}: Methodology, Validation, Writing, Funding acquisition.

\bibliography{main-bibliography}


\begin{appendices}
\section{Complexity Analysis}\label{Appendix:Complexity}
The computational overhead of our QCAD framework mainly comes from two parts: the calculation of Gower's Distance matrix using contextual features, and the calculation of anomaly score using all features based on Quantile Regression Forests. 

The time complexity of calculating Gower's Distance matrix is $O(N^{2}D_{cnt})$, where $N$ represents the sample size and $D_{cnt}$ denotes the number of contextual features. In addition, the time complexity of constructing a random forest is $O( n_{tree}\cdot n_{feature} \cdot k log(k))$ and using it for prediction is $O( n_{tree}\cdot n_{feature})$ \citep{buczak2015survey}, where $n_{tree}$ is the number of trees used to form a random forest, $n_{feature}$ denotes the number of dimensions (i.e., $D_{cnt}$ at most) and $k$ indicates the number of samples used to construct the forest (i.e., the number of nearest neighbours in our case). Hence, the time complexity of constructing $N$ quantile regression forests in $D_{bhv}$ behavioral features is  $O( n_{tree}\cdot D_{cnt} \cdot k log(k) \cdot N \cdot D_{bhv} )$. 
Moreover, we have to estimate $100$ different quantiles using each quantile regression forest, resulting in $O( (n_{tree} \cdot D_{cnt} \cdot k log(k)+ n_{tree} \cdot D_{cnt} \cdot 100) \cdot N \cdot D_{bhv} )$. Therefore, using our default setting leads to $O( ( 100 \cdot D_{cnt} \cdot \mathrm{min}(500,\frac{N}{2})\cdot \mathrm{log}(\mathrm{min}(500,\frac{N}{2})) + 100 \cdot D_{cnt} \cdot 100) \cdot N \cdot D_{bhv})$. In the worst case, it is $O(10^{5}N D_{cnt}D_{bhv}).$

Overall, the time complexity of our method is $O(10^{5}N D_{cnt}D_{bhv}+N^{2}D_{cnt})$ in the worst case. Particularly, the time complexity becomes $O(10^{5}N D_{cnt}D_{bhv})$ when $N<10^{5}$ (i.e., for small and medium datasets). Besides, the construction of quantile regression forests for each object in each behavioral feature can easily be performed in a parallel way, thus reducing the time complexity to $O(10^{5}N D_{cnt}+N^{2}D_{cnt})$ in the worst case.

\section{Dataset Description}\label{Appendix:Data}

\paragraph{Abalone} The dataset contains 4177 abalone physical measurement records. Specifically, we use the features Sex, Length, Diameter and Height as contextual features. Accordingly, we use the features Whole Weight, Shucked Weight, Viscera Weight, Shell Weight and Rings as behavioral features. This dataset is downloaded from UCI, \footnote{https://archive.ics.uci.edu/ml/datasets/abalone} containing no missing values.

\paragraph{Airfoil Self-Noise} This dataset contains 1503 records from a series of aerodynamic and acoustic tests of airfoil blade sections. Specifically, we use the features describing the frequency, the angle of attack, the chord length, the free-stream velocity and the suction side displacement thickness (e.g., f, alpha, c, U\_infinity and delta) as contextual features. Meanwhile, the feature describing the scaled sound pressure level (e.g., SSPL) as behavioral feature. This dataset is downloaded from UCI, \footnote{https://archive.ics.uci.edu/ml/datasets/Airfoil+Self-Noise} containing no missing values.

\paragraph{Bodyfat} This dataset contains the estimates of body density and 
the percentage of body fat (used as behavioral features), which are determined by various body circumference measurements, such as, Age, Weight, Height, Neck circumference, Chest circumference, Abdomen 2 circumference, Hip circumference, Thigh circumference, Knee circumference, Ankle circumference, Biceps (extended) circumference, Forearm circumference, and Wrist circumference (used as contextual features) for 252 men. The raw dataset includes no categorical features and 0 missing values, downloaded from the CMU statlib. \footnote{http://lib.stat.cmu.edu/datasets/bodyfat}

\paragraph{Boston House Price} This dataset contains 583 records of the Boston house price. We use the features describing the properties of the house and its surroundings, i.e., CRIM, ZN, INDUS, CHAS, NOX, RM, AGE, DIS, RAD, TAX, PTRATIO, B, LSTAT as contextual features, and the median value of owner-occupied homes, i.e., MED, as behavioral feature. This dataset is released by \cite{harrison1978hedonic} and downloaded from the CMU statlib \footnote{http://lib.stat.cmu.edu/datasets}. It remains 506 records after dropping rows containing missing values.

\paragraph{Concrete Compressive Strength } This dataset contains 1030 records about the compressive strength of concrete. We use the features describing the age and different ingredients as contextual features, including C1, C2, C3, C4, C5, C6, C7 and Age. In addition, we use the feature Strength as behavioral feature. This dataset is downloaded from UCI, \footnote{https://archive.ics.uci.edu/ml/datasets/Concrete+Compressive+Strength} containing no missing values.

\paragraph{El Nino } This dataset contains 178080 records of the oceanographic and surface meteorological readings, which are 
taken from buoys located throughout the equatorial Pacific. We use the temporal and spatial features, i.e., Year, Month, Day, Date, Latitude, Longitude as contextual features, and other features, i.e., Zonal\_Winds, Meridional\_Winds, Humidity, Air\_Temp, Sea\_Surface\_Temp as behavioral features. This dataset is downloaded from the UCI machine learning repository.\footnote{archive.ics.uci.edu/ml/datasets/El+Nino} It remains 93935 records after dropping rows containing missing values. However, in order for all algorithms to complete the experiment on this dataset within 12 hours, we randomly downsample the original dataset to 20,000 records.

\paragraph{Energy Efficiency} This dataset contains 768 records about a study which assesses the energy efficiency as a function of building parameters. Accordingly, the building parameters such as X1, X2, X3, X4, X5, X6, X7 and X8 are regarded as contextual features, and the heating load and cooling load (namely Y1 and Y2) are treated as behavioral features. This dataset is downloaded from UCI, \footnote{https://archive.ics.uci.edu/ml/datasets/Energy+efficiency} containing no missing values.

\paragraph{Fish Weight } This dataset contains 157 records about the features of common species of fish in the market. Accordingly, we use the features including Species, Length1, Length2, Length3, Height, Width as contextual features, and Weight as behavioral feature. This dataset is downloaded from Kaggle, \footnote{https://www.kaggle.com/aungpyaeap/fish-market} containing no missing values.

\paragraph{Forest Fires} This dataset contains 517 records concerning meteorological and spatiotemporal information about forest fires in the northeast region of Portugal. We use the spatiotemporal features such as  X, Y, month, day as contextual features, and the rest features, i.e., FFMC, DMC, DC, ISI, temp, RH, wind, rain, area as behavioral features. It is downloaded from the UCI machine learning repository,\footnote{https://archive.ics.uci.edu/ml/datasets/forest+fires} containing no missing values.

\paragraph{Gas Turbine CO and NOx Emission } The original dataset includes 36733 records of sensor measures data, which is collected from a gas turbine in Turkey from 2011 to 2015. The dataset is collected from the same power plant to study flue gas emissions and predict the hourly net energy yield. Consequently, we use the features describing the turbine parameters (e.g., AT, AP, AH, AFDP, GTEP, TIT, TAT and CDP) as contextual features.
The variables characterizing the turbine energy yield and emissions of gas  (e.g., TEY, CO and NOX) are used as behavioral features. However, for all algorithms to complete the experiments within 12 hours, we only use the 7383 records in 2015. This dataset is downloaded from UCI, \footnote{https://archive.ics.uci.edu/ml/datasets/Gas+Turbine+CO+and+NOx+Emission+Data+Set} containing no missing values.

\paragraph{Heart Failure} 
This dataset contains heart failure clinical records of 299 patients suffering from heart failure. It contains 13 clinical features, of which we use age, sex, smoking, diabetes, high\_blood\_pressure, and anaemia as contextual features, and creatinine\_phosphokinase, ejection\_fraction, platelets, serum\_creatinine, serum\_sodium, time as behavioral features. Besides, the death\_event feature is removed and there is no missing value in this dataset. The original dataset is released by \cite{ahmad2017survival} and we download it from the UCI machine learning repository. \footnote{archive.ics.uci.edu/ml/datasets/Heart+failure+clinical+records}

\paragraph{Hepatitis}
This dataset contains 615 records concerning the laboratory values of blood donors and Hepatitis C patients. We use the demographic features such as sex and age, and Category of donors as contextual features, and the rest of features, i.e., ALB, ALP, ALT, AST, BIL, CHE, CHOL, CREA, GGT and PROT as behavioral features. The original dataset is downloaded from the UCI machine learning Repository. \footnote{ https://archive.ics.uci.edu/ml/datasets/HCV+data} 26 records contain missing values and therefore are removed.

\paragraph{Indian Liver Patient} 
This dataset contains the medical records of 583 Indian liver patients, 4 of which contain missing values and are therefore deleted from further analysis. We treat Age, Gender and Selector as contextual features, and Total\_Bilirubin, Direct\_Bilirubin, Alkaline\_Phosphotase, Alamine\_Aminotransferase, Aspartate\_Aminotransferase, Total\_Protiens, Albumin, Albumin\_and\_Globulin\_Ratio as behavioral features. This dataset is downloaded from the UCI machine learning repository. \footnote{https://archive.ics.uci.edu/ml/datasets/ILPD+(Indian+Liver+Patient+Dataset)}

\paragraph{Maintenance of Naval Propulsion Plants } The data set contains 11934 experimental records, which were performed by a numerical simulator of a naval vessel featuring a gas turbine propulsion plant. We use the features describing the gas turbine measures of the physical asset as contextual features, including LeverPosition, GTT, GTn, GGn, Ts, Tp, T48, T1, T2, P48, P1, P2, Pexh, TIC, mf. Meanwhile, we use the features containing the ship speed and the performance decay over time of gas turbine components, e.g., ShipSpeed, CompressorDecay and TurbineDecay, as behavioral features. This dataset is downloaded from UCI, \footnote{https://archive.ics.uci.edu/ml/datasets/Condition+Based+Maintenance+of+Naval+Propulsion+Plants} containing no missing values.

\paragraph{Parkinsons Telemonitoring} The data set contains 5875 records of a series of biomedical voice measurements from 42 early-stage Parkinson's disease patients. There people were recruited into a six-month trial of remote monitoring equipment for remote symptom progression monitoring. The features such as subject, age, sex, test\_time, Jitter, Jitter\_Abs, Jitter\_RAP, Jitter\_PPQ5, Jitter\_DDP, Shimmer, Shimmer\_dB, Shimmer\_APQ3, Shimmer\_APQ5, Shimmer\_APQ11, Shimmer\_DDA, NHR, HNR, RPDE, DFA, PPE descirbe the background information, and thus are used as contextual features. Meanwhile, the corresponding scores motor\_UPDRS and total\_UPDRS are used as behavioral features. This dataset is downloaded from UCI, \footnote{https://archive.ics.uci.edu/ml/datasets/Parkinsons+Telemonitoring} containing no missing values.

\paragraph{Power Plant} This dataset contains 9568 records from a combined cycle power plant which worked with full load from 2006 to 2011. Features such as hourly average ambient variables temperature, ambient pressure, relative humidity and exhaust vacuum (e.g., T, AP, RH and EP) are considered as contextual features, while the net hourly electrical energy output (EP) is regarded as behavioral feature. This dataset is downloaded from UCI, \footnote{https://archive.ics.uci.edu/ml/datasets/Combined+Cycle+Power+Plant} containing no missing values.

\paragraph{QS University Ranking} This dataset contains the QS rankings of the world universities from 2018 to 2020. We consider the wolrd rankings, national rankings and location of each university from 2018 to 2020, i.e., World2018, National2018, World2019, National2019,World2020, National2020, Country as contextual features. We use six features that are considered for the ranking (i.e., Academic Reputation, Employer Reputation, Faculty to Student Ratio, Number of citations per faculty, International Faculty, International Students) as behavioral features. The original data is crawled from the QS website, \footnote{https://www.topuniversities.com/qs-world-university-rankings} containing 475 records after dropping missing values.

\paragraph{QSAR Fish Toxicity} This dataset contains 908 records about the fish Pimephales promelas. Specifically, the six features describing the molecular of 908 chemicals will be used as contextual features, and the corresponding acute aquatic toxicity measure will be used as behavioral feature. This dataset is downloaded from UCI, \footnote{https://archive.ics.uci.edu/ml/datasets/QSAR+fish+toxicity} containing no missing values.

\paragraph{Synchronous Machine } This dataset contains 557 records from a real experimental set. This experiment aims to construct a model to estimate the excitation current of synchronous motors. Specifically, we use the features describing the load current, power factor, power factor error and changing of excitation current (e.g., Iy, PF, e and dIf) as contextual features. Accordingly, the feature which describes
the excitation current of synchronous machine (namely If ) is used as behavioral feature. This dataset is downloaded from UCI, \footnote{https://archive.ics.uci.edu/ml/datasets/Synchronous+Machine+Data+Set} containing no missing values.

\paragraph{Yacht Hydrodynamics } This dataset contains 308 records about the features of sailing yachts. We use the features describing the dimensions, velocity and hydrodynamic performance of yachts, namely Longitudinal\_position, Prismatic\_coefficient, Length\_displacement\_ratio, Beam\_draught\_ratio, Length\_beam\_ratio, Froude\_number as contextual features, and the feature concerning the residuary resistance per unit weight of displacement, namely resistance, as behavioral feature. This dataset is downloaded from the UCI machine learning repository, \footnote{archive.ics.uci.edu/ml/datasets/Yacht+Hydrodynamics} containing no missing values.

\section{Parameter Sensitivity Analysis}
\label{Appendix:Sensitivity}

The parameter $k$ directly affects the effectiveness and efficiency of anomaly detector in anomaly detection phase. If our dataset is labelled, we can use techniques like grid-search and cross-validation to set an optimal $k$ for a specific dataset. However, anomaly detection is usually an unsupervised learning problem, which makes it impossible to set an optimal $k$. Hence, we can only empirically give some rules of thumb to set $k$ according to the properties of specific dataset (i.e., sample size, dimensionality and estimated rate of anomaly). Therefore, we perform intensive experiments on synthetic and real-world datasets with a wide range of sample sizes, dimensionalities and rates of injected anomalies to investigate the sensitivity of our algorithm on this parameter. 

\begin{figure}[!htbp]
\centering
\includegraphics[width=12cm]{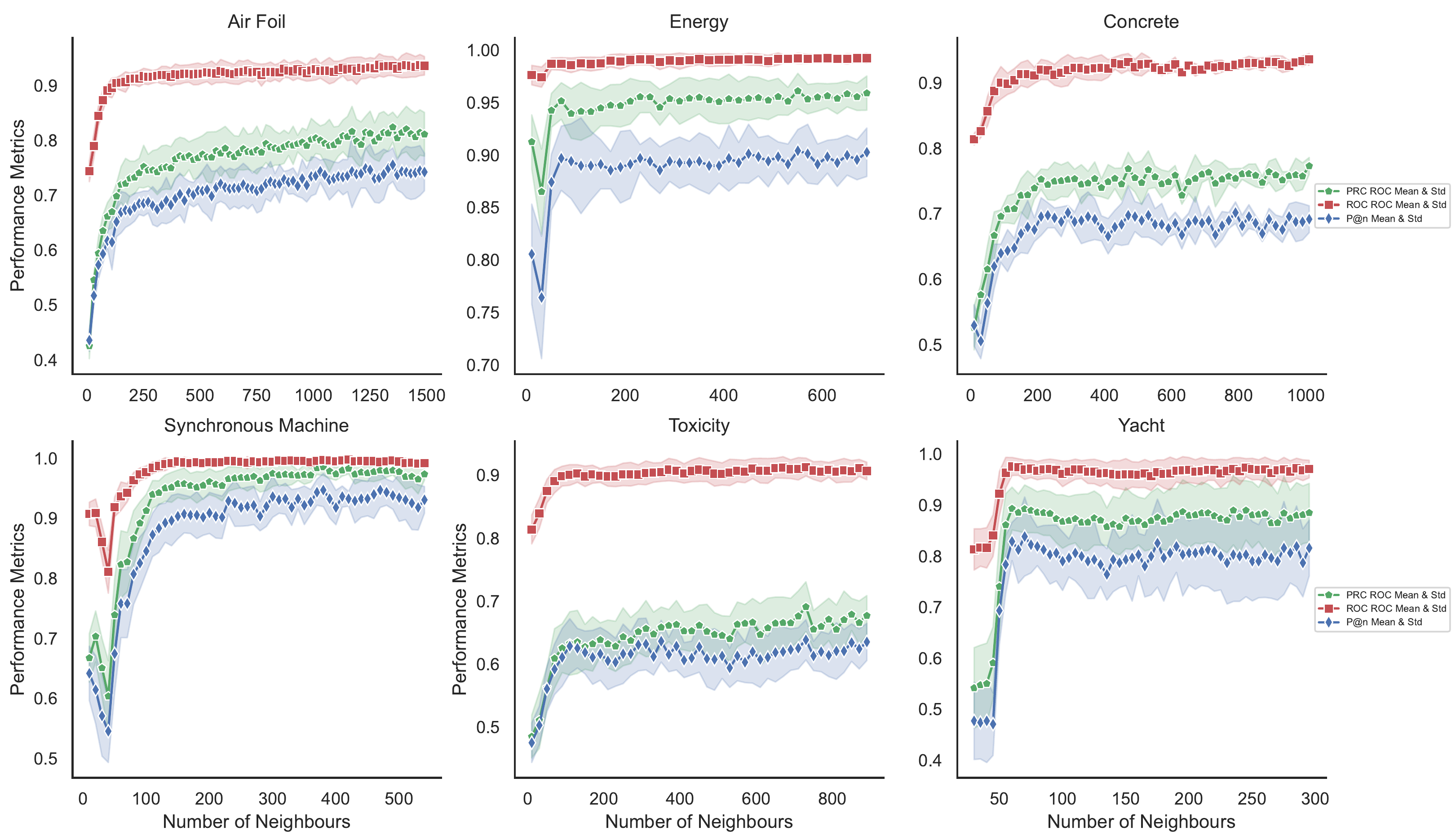}
\caption{Sensitivity analysis on parameter $k$, where lines represent the mean values and the shaded areas indicate the corresponding standard deviations of each metric on 10 independent trials. Increasing the number of neighbours will first largely improve the performance in terms of PRC AUC (green line with asterisks), ROC AUC (red line with squares) and Precision@n (blue line with diamonds), and then yields negligible performance gains. }\label{fig:sensitivity1}
\end{figure}

As shown in Figure \ref{fig:sensitivity1}, increasing the 
number of neighbours, i.e., $k$,  will lead to a better performance in terms of PRC AUC, ROC AUC and Precision@n when $k$ is small (about $N/10$ for most datasets). However, the performance gain gradually slows down as $k$ increases, and the performance finally reaches a plateau with an increase of $k$. After that, further increasing $k$ yields only a negligible performance gain, at the cost of runtime. Therefore, we set $k$ to $N/2$ for small dataset and $500$ for medium dataset after taking a trade-off between the accuracy and runtime cost. Overall, different from other $k$-NN based anomaly detectors which are sensitive to this parameter, our method QCAD is stable on parameter $k$ as long as its value is not too small.

\section{Ablation Study}
\label{Appendix:Ablation}
\subsection{Scaling Conditional Quantile Interval Length}

\begin{figure}[!htbp]
\centering
\includegraphics[width=12cm]{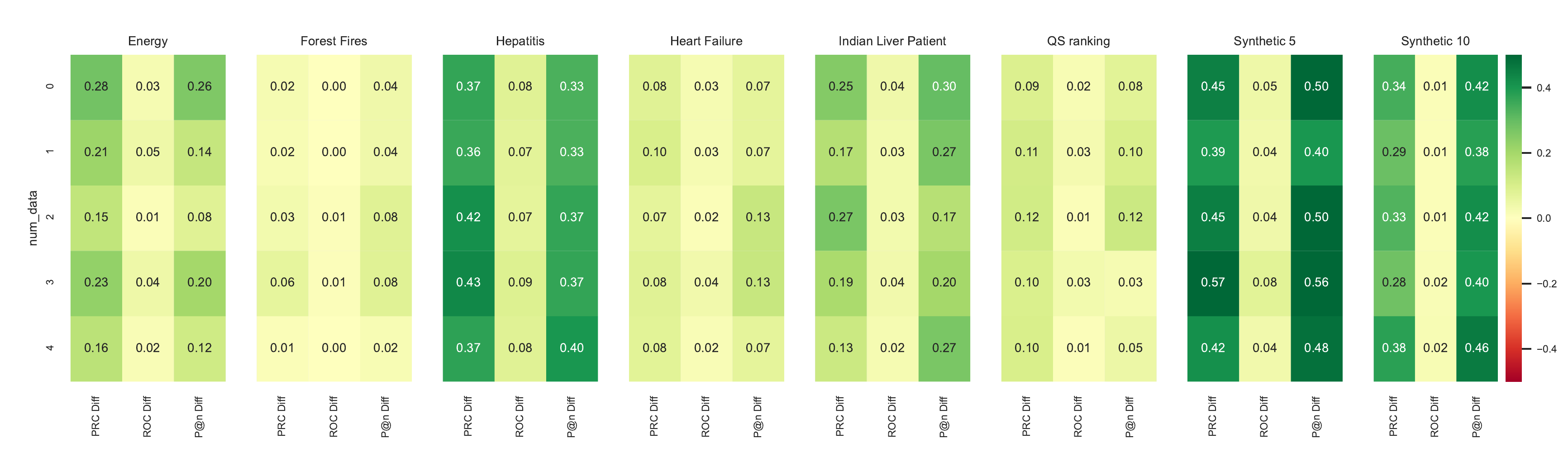}
\caption{Effects of scaling matched conditional quantile interval length. For each dataset, the results are obtained by performing 5 independent trials of injecting contextual anomalies. For each trial, which is represented by the y-axis (namely $num\_data$), the results are the difference (in terms of PRC AUC, ROC AUC, Precision@n, respectively) of methods with or without scaling the matched conditional quantile interval length. The large differences on most datasets imply the critical importance of scaling matched conditional quantile interval length. }\label{fig:ablation1}
\end{figure}

In our method section, we have defined a matched conditional quantile interval length for an observation $b^{q}$ when $b^{q} > \tau_{100}^{q}$ or $ b^{q} <\tau_{0}^{q}$ as in Equation (4). We define the matched interval length for $b^{q}$ based on $\mathrm{max}(w(\mathbf{x}\rvert\mathbf{b}^{q}))$ and further scale it by considering its distance to $\tau_{100}^{q}$ or $\tau_{0}^{q}$. Alternatively, we can simply define the matched interval length as $\mathrm{max}(w(\mathbf{x}\rvert\mathbf{b}^{q}))$ without scaling it. In other words, we could treat all observations residing outside $[\tau_{0}^{q}, \tau_{100}^{q}]$ equally. However, as shown in Figure \ref{fig:ablation1}, this will lead to a large performance degradation  in terms of PRC AUC and Precision@n for most datasets. 

Concretely, Figure \ref{fig:ablation1} shows the difference of performance metrics, i.e., PRC ROC, ROC AUC and Precision@n  between scaling $\mathrm{max}(w(\mathbf{x}\rvert\mathbf{b}^{q}))$ with considering the distance to $\tau_{100}^{q}$ or $\tau_{0}^{q}$, and not scaling $\mathrm{max}(w(\mathbf{x}\rvert\mathbf{b}^{q}))$. For all datasets, the differences are positive. 
Particularly, these differences are significantly large in terms of PRC AUC and Precision@n for most datasets such as Energy, Hepatitis, Indian Liver Patient, Synthetic 5 and  Synthetic 10. Therefore, it is pivotal to define the matched interval length by considering the distance from $b^{q}$ to $\tau_{100}^{q}$ or $\tau_{0}^{q}$ when $b^{q}> \tau_{100}^{q}$ or $b^{q}<\tau_{0}^{q}$, respectively.

\subsection{Clipping Conditional Quantile Interval Length}
\begin{figure}[!htbp]
\centering
\includegraphics[width=12cm]{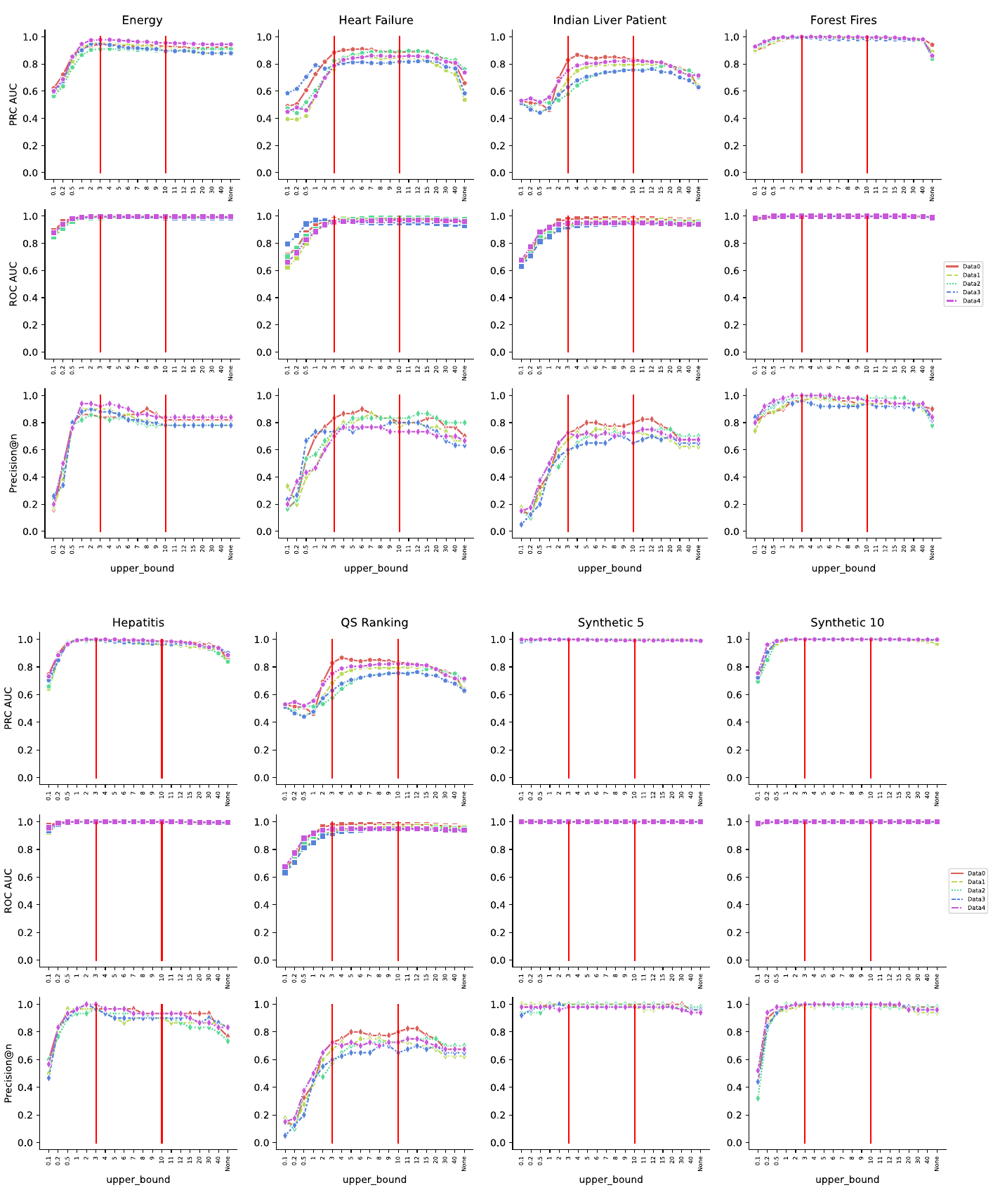}
\caption{Effects of clipping matched conditional quantile interval length. For each dataset, the results are obtained by performing 5 independent trials (represented by 5 different curves) of injecting contextual anomalies. Then we compute the performance metrics (i.e., PRC AUC, ROC AUC, and Precision@n) of QCAD by varying the hyper-parameter $\eta$ in $\{0.1,0.2,0.5,1,2,3,...,11,12,15,20,30,40,None\}$, where \textit{None} means no clipping is performed. On most datasets, the best performances are achieved when $3<\eta<10$, as shown by the two vertical red lines.}\label{fig:ablation2} 
\end{figure}

To mitigate \textit{dictator effect}, we have proposed to clip large matched conditional quantile interval lengths with an upper bound, which is set to $\frac{\eta}{100}$. More concretely, we set $\eta=10$. To demonstrate the efficacy of this strategy on various datasets, we compute the performance metrics of QCAD by varying the hyper-parameter $\eta \in \{0.1,0.2,0.5,1,2,3,...,11,12,15,20,30,40,None\}$, where \textit{None} means no clipping is performed.

As shown in Figure \ref{fig:ablation2},  increasing $\eta$ will lead to  higher PRC AUC, ROC AUC and Precision@n values when $\eta$ is small (i.e., less than $1$ for most datasets). Next, on the one hand, the ROC AUC stays stable as $\eta$ increases, even without clipping (i.e. $\eta=None$). In other words, the clipping strategy does not affect ROC AUC metric. On the other hand, the PRC AUC and Precision@n gradually climb to a plateau as $\eta$ increases. After a certain period, the PRC AUC and Precision@n start decreasing with an increase of $\eta$. Overall, on most datasets, PRC AUC and Precision@n generally achieve higher values with $3\leq \eta \leq 10$ than those without clipping. Therefore, the \textit{dictator effects} indeed exist and they mainly 
reduce the performance of QCAD in terms of PRC AUC and Precision@n. Moreover, our proposed clipping strategy can effectively overcome this problem.  
\end{appendices}

\end{document}